\documentclass{article}

\usepackage[preprint]{neurips_2026}

\usepackage[utf8]{inputenc} 
\usepackage[T1]{fontenc}    
\usepackage{hyperref}       
\usepackage{url}            
\usepackage{booktabs}       
\usepackage{amsfonts}       
\usepackage{nicefrac}       
\usepackage{microtype}      
\usepackage{xcolor}         
\usepackage{multirow}
\usepackage{graphicx}
\usepackage{algorithm}
\usepackage{algorithmic}
\usepackage{amsmath}
\usepackage{tabularx}
\usepackage{array}
\usepackage{makecell}
\usepackage{caption}
\usepackage{rotating}
\setcitestyle{numbers,square,comma}

\title{MetaColloc: Optimization-Free PDE Solving via Meta-Learned Basis Functions}

\author{%
  Zichuan Yang \\
  School of Mathematical Sciences\\
  Tongji University\\
  Shanghai, P.R. China \\
  \texttt{2153747@tongji.edu.cn} \\
}

\begin{document}

\maketitle

\begin{abstract}
Solving partial differential equations (PDEs) with machine learning typically requires training a new neural network for every new equation. This optimization is slow. We introduce MetaColloc. It is an optimization-free and data-free framework that removes this bottleneck completely. We decouple basis discovery from the solving process. We meta-train a dual-branch neural network on diverse Gaussian Random Fields. This offline process creates a universal dictionary of neural basis functions. At test time, we freeze the network. We solve the PDE by assembling a collocation matrix. We find the solution through a single linear least squares step. For non-linear PDEs, we apply the Newton-Raphson method to achieve fast quadratic convergence. Our experiments across six 2D and 3D PDEs show massive improvements. MetaColloc reaches state-of-the-art accuracy on smooth and non-linear problems. It also reduces test-time computation by several orders of magnitude. Finally, we provide a detailed frequency sweep analysis. This analysis reveals a critical mismatch between function approximation and operator stability at extremely high frequencies. This profound finding opens a clear path toward future operator-aware meta-learning.
\end{abstract}

\section{Introduction}
\label{sec:1}
Partial differential equations (PDEs) describe many physical systems. They appear in fluid flow, quantum models, and electromagnetic fields. Classical solvers work well but require complex meshes. Their cost grows fast in high dimensions. Recent machine learning methods try to avoid this cost. Physics-Informed Neural Networks (PINNs) \cite{raissi2019physics} turn PDE solving into an optimization problem. They use neural networks and automatic differentiation to match the PDE residual. This idea removes the mesh and works in high-dimensional spaces.

PINNs still face a core problem. They must train a new network for every new PDE. This training takes thousands of gradient steps. It is slow and often unstable. Neural networks also show spectral bias \cite{rahaman2019spectral}. They learn low frequencies first and struggle with high-frequency waves. They also face challenges with stiff non-linear equations. Other methods try to speed up test-time solving. Fourier Neural Operators (FNO) \cite{li2020fourier} and DeepONets \cite{lu2021learning} learn mappings between function spaces. They solve new PDEs quickly. But they need large paired datasets. These datasets come from classical solvers and are expensive to generate.

We take a different view. We look at the structure of PDE solvers. The hard part is not the final solve. The hard part is finding a good set of basis functions. If we have a strong basis, the PDE becomes a simple linear least-squares problem. This solve takes only a few seconds. This idea suggests a shift. We should learn the basis offline. We should keep the online stage simple and fast.

Some earlier works try this idea. Extreme Learning Machines (ELMs) \cite{dong2021local} and random feature models use random networks as basis functions. These methods are simple but limited. To approximate complex functions, random features often require a very large number to achieve effective learning outcomes \cite{gavrikov2023powerlinearcombinationslearning}. Furthermore, they are highly sensitive to initialization.

We propose MetaColloc to address these limits. MetaColloc is an optimization-free PDE solver, requiring neither solution data nor gradient-based optimization at test time. We use meta-learning to build a universal dictionary of neural basis functions. We train the network offline on many Gaussian Random Field (GRF) tasks. GRFs are natural priors for PDE solutions. They cover smooth and rough functions. This forces the network to learn a wide range of shapes. To handle extremely high frequencies, we use a dual-branch design. One branch processes raw coordinates. The other branch uses multi-scale Fourier features \cite{tancik2020fourier}. The network learns how to mix these two branches during meta-training.

\begin{figure}[h]
    \centering
    \includegraphics[width=\textwidth]{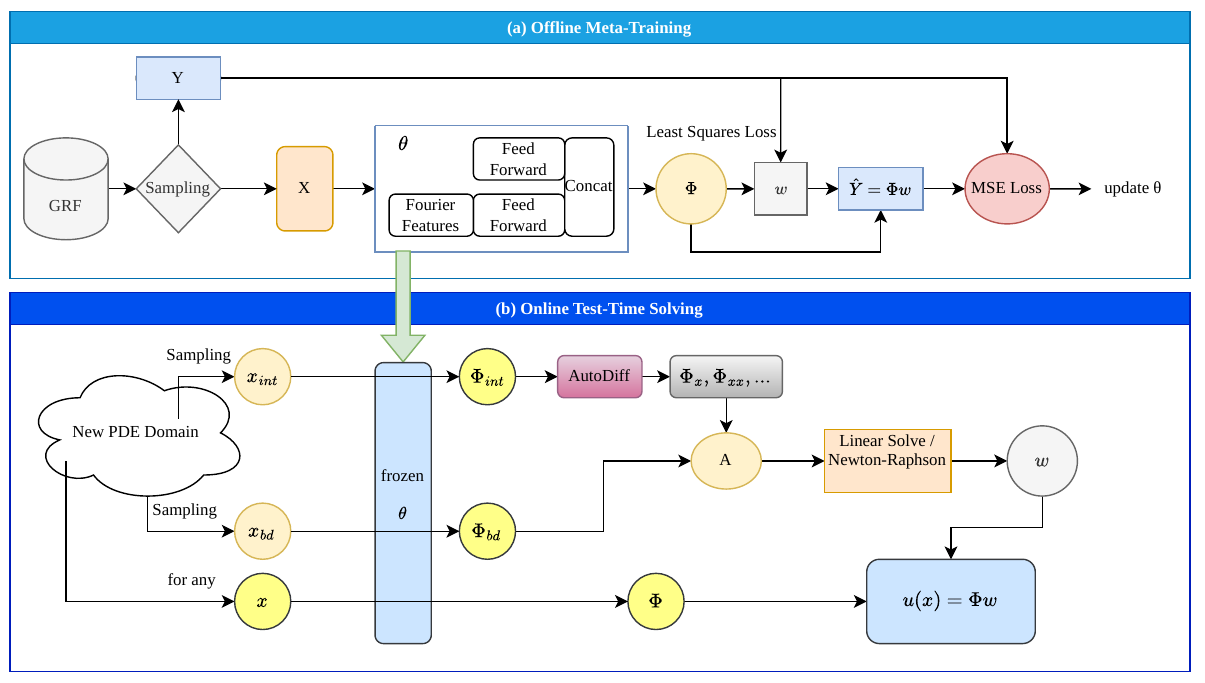}
    \caption{\textbf{Overview of MetaColloc.} During meta-training (a), coordinate samples $x$ are drawn from multi-scale Gaussian Random Fields (GRF). The dual-branch network processes inputs through a low-frequency branch (raw coordinates) and a high-frequency branch (multi-scale Fourier embedding), whose outputs are concatenated to form the basis matrix $\Phi$. Network parameters $\theta$ are updated via least-squares loss on GRF tasks. At test time (b), $\theta$ is frozen; given collocation points $x_{int}$ from a new PDE, the network produces $\Phi$ and its derivatives $\Phi_x, \Phi_{xx}$, ..., which are assembled into a linear system solved in closed form. The PDE solution is then recovered as $u(x) = \Phi(x) \cdot w$, requiring neither solution data nor gradient-based optimization.}
    \label{fig:metacolloc}
\end{figure}

Figure~\ref{fig:metacolloc} shows the full pipeline. At test time, we freeze the network. We compute exact derivatives with forward-mode automatic differentiation. We build a collocation matrix and solve it in one step. For non-linear PDEs, we apply the Newton--Raphson method. This gives fast and stable updates. We reach high accuracy in a few iterations. We test MetaColloc on six PDEs. It matches state-of-the-art accuracy on smooth problems. It outperforms existing methods on high-frequency and non-linear equations. It also reduces test-time cost by several orders of magnitude.

\section{Related Work}
\label{sec:2}
\subsection{Physics-Informed Neural Networks}
Physics-Informed Neural Networks (PINNs) \cite{raissi2019physics} encode PDE constraints directly into a loss function. They solve equations using gradient descent. Many studies improve this baseline. Researchers balance loss terms during training \cite{wang2020understanding, DCGD2024, ConFIG2025}, analyze gradients with Neural Tangent Kernels \cite{wang2021eigenvector}, and enforce causality \cite{wang2022respecting}. Other works design new optimization techniques \cite{KFAC2024, Metamizer2025, RoPINN2024, PRDP2025} or propose new architectures like Kolmogorov-Arnold Networks \cite{KAN2025}. To solve high-frequency problems, some methods use random Fourier features \cite{wang2021eigenvector} or dynamic meshes \cite{PIG2025}. However, these methods remain very sensitive to hyper-parameters. 

All these approaches share a core limitation. They perform per-instance optimization. A user must train a new network for every new PDE. This makes test-time computation very expensive. MetaColloc avoids this problem completely. We move the heavy learning process to an offline meta-training stage. At test time, MetaColloc only requires a single linear solve.

\subsection{Gaussian Process and Generative PDE Solvers}
Gaussian Process (GP) methods provide a probabilistic way to solve PDEs \cite{chen2021solving}. Recent work like GP-HM \cite{fang2024gaussian} models the power spectrum of the solution directly in the frequency domain. GP-HM beats standard PINNs on high-frequency PDEs. Alongside GPs, recent years have shown a surge in diffusion models for PDE modeling. Researchers use generative diffusion to reconstruct physical fields, control complex systems, and predict dynamics \cite{DiffusionPDE2024, ConditionalDiffusion2024, Text2PDE2025, PhysicsInformedDiffusion2025, CLDiffPhyCon2025}. 

We differ from these methods fundamentally. GP-HM is still a per-instance solver. It optimizes kernel parameters for hundreds of seconds on every new problem. It also requires data points on a strict Cartesian grid. Generative diffusion models often require slow, iterative sampling steps \cite{TSM2025}. MetaColloc has no grid limits. Our test-time speed is extremely fast and remains independent of the PDE type.

\subsection{Operator Learning and Meta-Learning for PDEs}
Many researchers use operator learning to accelerate PDE simulations. Fourier Neural Operators (FNO) \cite{li2020fourier} and DeepONets \cite{lu2021learning} learn mappings between function spaces. Recent advances scale these operators heavily. They use Transformers \cite{AROMA2024, UPT2024, Poseidon2024, CViT2025}, State-Space Models \cite{MambaNO2024}, and geometry-aware deformations \cite{GeoFNO2024, LNO2024}. Some studies explore multi-physics pre-training \cite{MPP2024, CoDANO2024} and integrate Newton methods for non-linear operators \cite{NINO2024}. 

These neural operators generalize well across different physical parameters. However, they need massive datasets of exact PDE solutions. Generating this training data takes enormous computational power. MetaColloc trains completely without physical data. We sample random functions from Gaussian Random Fields to train our model. Furthermore, we do not learn a specific operator mapping. We learn a universal dictionary of basis functions.

\subsection{Random Features and Kernel Methods}
Random Fourier features \cite{rahimi2007random} efficiently approximate kernel functions. Extreme Learning Machines (ELM) \cite{huang2006extreme, dong2021local} use fixed, random hidden layers to solve PDEs. They turn the PDE problem into a linear system. Some recent works integrate random networks with time-parallel solvers \cite{RandNetParareal2024} or high-dimensional graph models \cite{SINGER2025}.

While ELM and random feature methods can solve simple PDEs with remarkable speed, their performance in collocation systems often exhibits high variance, being sensitive to the random initialization of the basis function shape parameters. For complex PDEs, this randomization strategy may require extensive hyperparameter tuning across GRFs with different correlation lengths, typically by selecting the shape parameter that minimizes the average MSE over these GRFs \cite{zhang2024transferable}. Alternatively, one can use an empirical equation-specific formula to adjust the shape parameter \cite{LU2025113902} for more stable representations, although its performance still depends on initialization and random seeds. In contrast, MetaColloc aims to systematically structure this representational power. We replace unoptimized random guessing with a deeply optimized dictionary shaped via meta-learning over multi-scale GRFs. MetaColloc retains the closed-form speed of a linear solver, but utilizes meta-training to provide a robust, fixed dictionary designed to be expressive across diverse physical regimes out-of-the-box.

\section{The MetaColloc Framework}
\label{sec:3}
\subsection{Decoupling Basis Discovery from PDE Solving}
\label{sec:3.1}

Standard machine learning solvers, such as Physics-Informed Neural Networks (PINNs), typically formulate PDE solving as an instance-specific optimization problem. While this approach has been highly successful across various physical domains, it generally requires training a neural network via gradient descent for each new equation. This per-instance training can be computationally intensive and sensitive to optimization dynamics at test time.

To alleviate this test-time bottleneck, we propose an alternative perspective: decoupling the PDE problem into an offline basis discovery stage and an online solving stage. We assume the solution $u(x)$ can be effectively approximated as a linear combination of basis functions. We write: 

\[
u(x) \approx \Phi(x) w.
\]

Here, $\Phi(x) \in \mathbb{R}^H$ represents a basis dictionary generated by a neural network with $H$ output units, and $w \in \mathbb{R}^H$ is the vector of unknown coefficients. 

Unlike conventional optimization-based solvers, MetaColloc strictly freezes the network parameters at test time, transforming the network into a universal spatial feature extractor. Consequently, the only unknown in our framework is the coefficient vector $w$. By treating the neural basis as a frozen dictionary, we bypass iterative backpropagation entirely, allowing $w$ to be efficiently determined through closed-form linear algebra operations.

\subsection{Implementation Details of the Dual-Branch Basis Dictionary}
\label{sec:3.2}

The core of MetaColloc is the neural network $\Phi_\theta: \mathbb{R}^d \to \mathbb{R}^H$, which parameterizes the universal basis dictionary. Here, $d$ is the spatial dimension (e.g., $d=2$ for 2D Poisson) and $H$ is the total number of basis functions. Crucially, to overcome the spectral bias \cite{rahaman2019spectral} inherent in standard MLPs and capture both smooth solutions and high-frequency oscillations without retraining, we adopt a decoupled dual-branch architecture. Let the input coordinate be $\mathbf{x} \in \mathbb{R}^d$. The network output $\boldsymbol{\phi}(\mathbf{x}) \in \mathbb{R}^H$ is composed as follows:

\[
\boldsymbol{\phi}(\mathbf{x}) = \text{Concat} \left[ \boldsymbol{\phi}_\text{low}(\mathbf{x}), \boldsymbol{\phi}_\text{high}(\mathbf{x}) \right]
\]

where $\boldsymbol{\phi}_\text{low} \in \mathbb{R}^{H/2}$ and $\boldsymbol{\phi}_\text{high} \in \mathbb{R}^{H/2}$ are outputs from the low- and high-frequency branches, respectively (assuming an equal split of the total basis budget $H$).

\paragraph{Low-Frequency Branch (Smooth Approximation):} This branch processes the raw input $\mathbf{x}$ to represent the macro-scale, smooth components of the physical field. It consists of a standard Multi-Layer Perceptron (MLP). To enhance gradient flow and representational capacity, we utilize the SwiGLU activation function \cite{chowdhery2022palmscalinglanguagemodeling}. A single SwiGLU layer $k$ with input $\mathbf{h}_{k-1}$ is formulated as:

\[
\mathbf{h}_k = \text{SwiGLU}(\mathbf{h}_{k-1}) = (\text{SiLU}(\mathbf{h}_{k-1} W_1 + \mathbf{b}_1)) \odot (\mathbf{h}_{k-1} W_2 + \mathbf{b}_2)
\]

where $\odot$ denotes element-wise multiplication, and SiLU($z$) = $z / (1 + \exp(-z))$. The low-frequency branch features 2 hidden SwiGLU layers, with widths of $H$ and $H/2$, respectively.

\paragraph{High-Frequency Branch (Spectral Enhancement):} To effectively model sharp gradients or oscillating solutions (e.g., the HighFreq Poission case in Section~\ref{sec:4.2}), this branch utilizes Multi-scale Fourier Features. The input $\mathbf{x}$ is first projected into a set of sinusoidal encodings:

\[
\boldsymbol{\gamma}(\mathbf{x}) = \left[ \sin(\pi \mathbf{k}_j^\top \mathbf{x}), \cos(\pi \mathbf{k}_j^\top \mathbf{x}) \right]_{j=1}^F
\]

Here, $F$ is the number of Fourier features, and the frequency vectors $\mathbf{k}_j \in \mathbb{R}^d$ are sampled from a multi-scale distribution to cover a wide spectrum. In our implementation, we explicitly set $\mathbf{k}_j$ as a set of fixed, axis-aligned frequency scales: $\{1.0, 2.0, 4.0, 8.0, 16.0, 32.0, 64.0, 128.0\}$, ensuring the dictionary possesses a priori representational power for high-frequency bands. The encoded features $\boldsymbol{\gamma}(\mathbf{x})$ are then passed through 2 SwiGLU hidden layers (with widths of $H$ and $H/2$, respectively) to form $\boldsymbol{\phi}_\text{high}(\mathbf{x})$.

\subsection{Meta-Training on Gaussian Random Fields}
\label{sec:3.3}
We want a neural network that provides a strong and flexible basis $\Phi(x)$. We use meta-learning to build this basis. The choice of the meta-training distribution matters. We use Gaussian Random Fields (GRFs). GRFs have strong links to Gaussian process theory \cite{williams2006gaussian}. Their reproducing kernel Hilbert spaces (RKHSs) have good approximation properties \cite{aronszajn1950theory,ha2006reproducing}. The RKHS of a GRF matches its Cameron--Martin space \cite{lukic2001stochastic}. This space describes the regularity and main directions of variation of the field.

Sample paths of GRFs do not lie in the RKHS. But their smoothness relates to Sobolev embeddings \cite{henderson2024sobolev}. This makes GRFs a natural prior for PDE solutions. They cover many function types. They also provide a principled way to sample functions for meta-training.

We sample GRFs at many scales. We vary the length scales across two orders of magnitude. We mix smooth radial kernels with oscillatory periodic kernels. This forces the network to learn smooth shapes and sharp waves. The basis then generalizes well to unseen PDEs.

We use a dual-branch network to capture these features. One branch takes raw coordinates. It focuses on smooth, low-frequency structure. The other branch takes multi-scale Fourier features \cite{tancik2020fourier}. It captures high-frequency oscillations. We use SwiGLU activations \cite{chowdhery2022palmscalinglanguagemodeling} in both branches. We concatenate their outputs to form $\Phi(x)$. We train the network to minimize least-squares error on many GRF tasks. Algorithm~\ref{alg:meta_train} shows this offline process.

\begin{algorithm}[ht]
\caption{Meta-Training the Neural Basis Dictionary}
\label{alg:meta_train}
\begin{algorithmic}[1]
\REQUIRE Meta-training epochs $E$, tasks per epoch $T$, neural network $\Phi_\theta$
\FOR{epoch $= 1$ to $E$}
    \FOR{task $= 1$ to $T$}
        \STATE Sample inputs $X$ and targets $Y$ from multi-scale GRF distribution
        \STATE Compute basis matrix $\Phi_\theta(X)$
        \STATE Solve for coefficients: $w = \text{lstsq}(\Phi_\theta(X), Y)$
        \STATE Predict outputs: $\hat{Y} = \Phi_\theta(X) w$
        \STATE Compute MSE loss between $\hat{Y}$ and $Y$
        \STATE Update $\theta$ with AdamW \cite{loshchilov2018decoupled}
    \ENDFOR
\ENDFOR
\RETURN Frozen network $\Phi_{\text{frozen}}$
\end{algorithmic}
\end{algorithm}

\subsection{optimization-free Collocation for Linear PDEs}
\label{sec:3.4}
After meta-training, the network $\Phi_{\text{frozen}}$ is ready for PDE solving. We use it inside a collocation framework. Consider the Poisson equation $-\Delta u = f$ with boundary condition $u = g$. We sample $N$ interior points and $N_b$ boundary points. We pass these points through the frozen network to get basis values.

We need derivatives of the basis. We use forward-mode automatic differentiation. It gives exact derivatives like $\Phi_{xx}$ and $\Phi_{yy}$ with low memory cost. We build the interior matrix by applying the PDE operator:

\[
-(\Phi_{xx} + \Phi_{yy}) w = f.
\]

We call this matrix $A_{eq}$. We build the boundary matrix

\[
\Phi_{bd} w = g.
\]

We call this matrix $A_{bd}$.

We stack $A_{eq}$ and $A_{bd}$. We also stack $f$ and $g$. This gives an overdetermined linear system. We solve it with least squares. This gives $w$. To evaluate the solution at a new point, we compute $\Phi_{\text{frozen}}(x)$ and multiply by $w$. We do not use gradient descent at test time.

\subsection{Newton--Raphson Iteration for Non-Linear PDEs}
\label{sec:3.5}
Many PDEs are non-linear. Examples include the Sine--Gordon and Korteweg--de Vries (KdV) equations. PINNs struggle with these problems. They must optimize a non-convex loss. They often face challenges on stiff systems.

We use a different idea. We apply the Newton--Raphson method in function space. We linearize the non-linear PDE operator iteratively using standard Newton-Raphson formulations. This turns the non-linear PDE into a short sequence of linear PDEs.

At each step, we evaluate the linearized operator with exact neural derivatives. We build a new least-squares system. We solve it to get an update $\Delta w$. We add this update to $w$. Each sub-problem is a convex linear system. This provides local quadratic convergence. We achieve high accuracy in five to eight iterations. Algorithm~\ref{alg:pde_solve} shows the full solver.

\begin{algorithm}[ht]
\caption{Universal PDE Solving (Linear and Non-Linear)}
\label{alg:pde_solve}
\begin{algorithmic}[1]
\REQUIRE Frozen network $\Phi_{\text{frozen}}$, PDE operator $\mathcal{L}$, boundary operator $\mathcal{B}$, iterations $K$
\STATE Sample interior points $X_{int}$ and boundary points $X_{bd}$
\STATE Compute basis matrices $\Phi_{int}$ and $\Phi_{bd}$
\STATE Compute derivatives (e.g., $\Phi_x$, $\Phi_{xx}$, ...) with forward-mode AD
\STATE Initialize $w = 0$
\STATE Set $K = 1$ for linear PDEs
\FOR{iteration $= 1$ to $K$}
    \STATE Compute predictions $u = \Phi w$ and needed derivatives
    \STATE Build interior matrix $A_{eq}$ from linearized $\mathcal{L}$
    \STATE Build boundary matrix $A_{bd}$ from linearized $\mathcal{B}$
    \STATE Form residuals for interior and boundary
    \STATE Stack $A = [A_{eq}; A_{bd}]$ and $R = [f; g]$
    \STATE Solve $\Delta w = \text{lstsq}(A, -R)$
    \STATE Update $w = w + \Delta w$
\ENDFOR
\RETURN Solution $u(x) = \Phi_{\text{frozen}}(x) w$
\end{algorithmic}
\end{algorithm}

\section{Experiments}
\label{sec:4}
We test MetaColloc on six different partial differential equations. We include linear, non-linear, smooth, and high-frequency problems. We place the detailed mathematical definitions of these equations and the hyperparameter settings in Appendix~\ref{app:a}. Additional numerical stability analysis—including floating-point precision studies and iteration-sweep convergence curves for nonlinear PDEs—are provided in Appendix~\ref{app:b.1}--\ref{app:b.2}. We also test our method on three-dimensional problems. We place those 3D results in Appendix~\ref{app:b.3}. Furthermore, we demonstrate that our frozen neural basis dictionary easily generalizes to complex irregular geometries (e.g., L-shape and Annulus domains) and mixed boundary conditions (Neumann and Robin) without any retraining. Detailed results are provided in Appendix~\ref{app:b.4}. 

\subsection{Ablation Study}
\label{sec:4.1}
We first study the design of our neural network. We want to understand why we need a dual-branch architecture and a meta-learning strategy. We compare our full MetaColloc model against four variations. The first variation uses only the low-frequency branch (We call it ``low-only'' in table). The second variation uses only the high-frequency branch (We call it ``high-only'' in table). The third variation is a non-learning model. This model mimics \citet{zhang2024transferable} approaches by using random weights and only tuning a single shape parameter. The fourth variation uses pure random features without any tuning. Table \ref{tab:rmse_pde_comparison_1} shows the results.

We observe a clear pattern in the results. The low-frequency branch works extremely well on smooth equations like the Poisson equation. However, it fails completely on the HighFreq problem. The high-frequency branch shows the opposite behavior. It struggles to achieve high precision on smooth problems but handles oscillations better. Our full MetaColloc model combines both branches. It achieves a strong balance and performs well across all tasks. 

Furthermore, the results prove that true meta-learning is necessary. The non-learning model and the random feature model perform very poorly. They cannot achieve high precision. Their random basis functions lack the true expressive power needed for complex physics. Our meta-training process successfully shapes the basis functions into a powerful dictionary.

\begin{table}[ht]
\centering
\small
\setlength{\tabcolsep}{3pt}
\caption{RMSE comparison across six PDEs. We report the mean and 95\% confidence interval. Lower is better.}
\label{tab:rmse_pde_comparison_1}
\begin{tabularx}{\textwidth}{l l X X X X X X}
\hline
\textbf{Method} & \textbf{Param} &
\textbf{Poisson} &
\textbf{Helmholtz} &
\textbf{VarCoeff} &
\textbf{HighFreq Poisson} &
\textbf{SineGordon} &
\textbf{KdV} \\
\hline

\multirow{4}{*}{MetaColloc}
& H=128  & 2.47e-3 (3.91e-4) & \textbf{5.00e-1 (1.74e-3)} & 1.79e-6 (7.29e-7) & 9.20e-1 (2.36e-1) & 1.56e-4 (5.32e-5) & 2.55e-3 (1.06e-3) \\
& H=256  & \textbf{4.04e-5 (1.19e-5)} & 5.02e-1 (1.90e-3) & 1.88e-9 (8.24e-10) & 7.81e-1 (8.24e-2) & 2.83e-7 (8.78e-8) & 1.80e-4 (5.20e-5) \\
& H=512  & 4.23e-5 (1.41e-6) & 5.08e-1 (2.48e-3) & 1.86e-9 (2.45e-10) & 8.26e-1 (1.19e-1) & 9.80e-8 (5.99e-8) & 1.59e-4 (4.88e-5) \\
& H=1024 & 4.28e-5 (1.33e-5) & 5.34e-1 (5.59e-3) & 1.93e-9 (5.86e-10) & 1.03e+0 (2.77e-1) & 6.03e-8 (2.97e-8) & 1.30e-4 (4.12e-5) \\
\hline

\multirow{4}{*}{high-only}
& H=128  & 5.09e-1 (1.52e-2) & 5.04e-1 (1.54e-3) & 5.32e-1 (2.22e-2) & 5.04e-1 (1.33e-2) & 3.52e-1 (3.28e-2) & 3.53e-1 (3.06e-2) \\
& H=256  & 5.06e-1 (7.71e-3) & 5.04e-1 (1.63e-3) & 5.26e-1 (4.01e-2) & 5.10e-1 (1.17e-2) & 3.19e-1 (7.41e-3) & 3.05e-1 (4.01e-2) \\
& H=512  & 5.04e-1 (2.64e-3) & 5.11e-1 (2.00e-3) & 5.26e-1 (2.91e-2) & 4.83e-1 (1.49e-2) & 2.85e-1 (1.17e-2) & 3.24e-1 (2.62e-2) \\
& H=1024 & 5.08e-1 (6.41e-3) & 5.90e-1 (1.26e-2) & 5.30e-1 (6.85e-3) & 4.79e-1 (6.54e-3) & 2.41e-1 (1.56e-2) & 2.99e-1 (2.65e-2) \\
\hline

\multirow{4}{*}{low-only}
& H=128  & 1.80e-4 (6.23e-5) & 5.04e-1 (1.89e-3) & 4.89e-9 (1.10e-9) & 2.49e+1 (7.30e+0) & 2.58e-8 (1.12e-8) & \textbf{2.69e-6 (7.07e-7)} \\
& H=256  & 4.33e-5 (2.11e-5) & 5.07e-1 (2.82e-3) & \textbf{1.29e-9 (3.29e-10)} & 5.04e+1 (4.36e+0) & \textbf{3.24e-9 (1.11e-9)} & 2.74e-6 (9.04e-7) \\
& H=512  & 2.55e-4 (2.23e-5) & 5.24e-1 (2.55e-3) & 4.35e-9 (3.66e-10) & 1.90e+2 (1.13e+1) & 5.29e-9 (2.69e-9) & 3.36e-6 (5.10e-7) \\
& H=1024 & 3.76e-4 (4.66e-5) & 5.64e-1 (6.27e-3) & 8.50e-9 (1.52e-9) & 3.37e+2 (1.72e+1) & 1.49e-8 (6.41e-9) & 3.37e-6 (7.11e-7) \\
\hline

\multirow{4}{*}{non-learning}
& H=128  & 1.23e-2 (5.01e-3) & 5.03e-1 (1.57e-3) & 1.80e-4 (2.37e-4) & 1.09e+0 (1.34e-1) & 4.08e-3 (5.73e-3) & 3.39e-2 (2.91e-2) \\
& H=256  & 1.48e-3 (1.21e-3) & 5.04e-1 (6.06e-4) & 2.38e-4 (2.92e-4) & 1.01e+0 (8.12e-2) & 1.17e-2 (1.33e-2) & 7.61e-2 (4.29e-2) \\
& H=512  & 9.19e-5 (5.77e-5) & 5.08e-1 (4.40e-3) & 4.52e-6 (4.46e-6) & 6.11e-1 (4.11e-1) & 2.50e-4 (2.34e-4) & 4.07e-2 (4.44e-2) \\
& H=1024 & 3.14e-3 (4.63e-3) & 5.28e-1 (1.03e-2) & 5.90e-6 (8.83e-6) & \textbf{4.69e-1 (5.04e-1)} & 1.72e-4 (2.58e-4) & 4.21e-2 (3.62e-2) \\
\hline

\multirow{4}{*}{RF-NLS \cite{liao2025solving}}
& N=128  & 3.07e-2 (9.90e-3) & 5.02e-1 (1.43e-3) & 4.38e-2 (1.33e-2) & 1.84e+1 (4.04e+0) & 1.24e-1 (6.34e-2) & 1.99e-1 (4.39e-2) \\
& N=256  & 2.73e-2 (4.33e-3) & 5.02e-1 (1.43e-3) & 3.73e-2 (3.80e-3) & 2.04e+1 (4.46e+0) & 8.86e-2 (4.57e-2) & 3.11e-1 (2.48e-2) \\
& N=512  & 2.23e-2 (7.98e-4) & 5.02e-1 (1.43e-3) & 2.99e-2 (2.14e-3) & 1.97e+1 (2.14e+0) & 6.04e-2 (3.77e-2) & 3.35e-1 (6.82e-2) \\
& N=1024 & 1.50e-2 (1.81e-3) & 5.02e-1 (1.43e-3) & 2.01e-2 (2.46e-3) & 1.90e+1 (2.44e+0) & 4.16e-2 (1.92e-2) & 2.90e-1 (2.53e-2) \\
\hline
\end{tabularx}
\end{table}

\subsection{Baseline Comparison}
\label{sec:4.2}
We compare MetaColloc against four strong baseline methods. The first is a standard PINN trained with the L-BFGS optimizer. The second is GP-HM, a state-of-the-art Gaussian Process method designed for high-frequency PDEs. The third is ConFIG, a recent method that balances training gradients. The fourth is a PINN combined with Dual Cone Gradient Descent (DCGD). Table \ref{tab:rmse_pde_comparison_2} shows the detailed numerical results. It is important to note that we do not include Neural Operators (such as FNO \cite{li2020fourier} or DeepONet \cite{lu2021learning}) in Table~\ref{tab:rmse_pde_comparison_2}. Neural Operators are strictly supervised operator learners that require massive, pre-computed paired datasets (generated by expensive classical solvers) to map between function spaces. In contrast, MetaColloc, PINNs, and GP-HM are data-free instance solvers. Comparing a data-free solver with methods trained on thousands of exact numerical solutions would be methodologically fundamentally unfair. Thus, we restrict our baseline comparisons to purely data-free, optimization-based instance solvers

MetaColloc shows a massive advantage on smooth and non-linear equations. Our method achieves errors on the order of $10^{-9}$ for the VarCoeff problem. The traditional PINN method only reaches $10^{-3}$. We see similar dominant results on the Sine-Gordon and KdV equations. MetaColloc handles strong non-linearities and high-order derivatives easily because our Newton-Raphson approach provides exact functional linearizations.

We must also discuss the extremely high-frequency Helmholtz problem. This problem acts as a disaster plateau for almost all methods. Traditional networks fail completely and yield an error of around 0.5. Only the GP-HM method solves this specific equation well. However, GP-HM is extremely slow. It takes roughly 4500 seconds to optimize a single equation. It also requires a dense uniform grid of 10000 points. In contrast, MetaColloc takes only one fourth at most, including the entire offline training phase. Crucially, at test-time, MetaColloc solves new instances in ~1.3 seconds (see Appendix~\ref{app:b.5}). Our solver also uses only 3200 random points. We also note that ConFIG and DCGD show better accuracy on the HighFreq problem. We accept this limitation. MetaColloc trades some extremely high-frequency accuracy for massive speed gains, data-free training, and perfect stability in non-linear physics.

\begin{table}[ht]
\centering
\small
\setlength{\tabcolsep}{3pt}
\caption{RMSE comparison against baseline methods. We report the mean and 95\% confidence interval. Lower is better.}
\label{tab:rmse_pde_comparison_2}
\begin{tabularx}{\textwidth}{l l X X X X X X}
\hline
\textbf{Method} & \textbf{Param} &
\textbf{Poisson} &
\textbf{Helmholtz} &
\textbf{VarCoeff} &
\textbf{HighFreq Poisson} &
\textbf{SineGordon} &
\textbf{KdV} \\
\hline

\multirow{4}{*}{MetaColloc (Ours)}
& H=128  & 2.47e-3 (3.91e-4) & 5.00e-1 (1.74e-3) & 1.79e-6 (7.29e-7) & 9.20e-1 (2.36e-1) & 1.56e-4 (5.32e-5) & 2.55e-3 (1.06e-3) \\
& H=256  & \textbf{4.04e-5 (1.19e-5)} & 5.02e-1 (1.90e-3) & 1.88e-9 (8.24e-10) & 7.81e-1 (8.24e-2) & 2.83e-7 (8.78e-8) & 1.80e-4 (5.20e-5) \\
& H=512  & 4.23e-5 (1.41e-6) & 5.08e-1 (2.48e-3) & \textbf{1.86e-9 (2.45e-10)} & 8.26e-1 (1.19e-1) & 9.80e-8 (5.99e-8) & 1.59e-4 (4.88e-5) \\
& H=1024 & 4.28e-5 (1.33e-5) & 5.34e-1 (5.59e-3) & 1.93e-9 (5.86e-10) & 1.03e+0 (2.77e-1) & \textbf{6.03e-8 (2.97e-8)} & 1.30e-4 (4.12e-5) \\
\hline

\multirow{4}{*}{PINN L-BFGS \cite{raissi2019physics}}
& H=128  & 3.61e-2 (3.44e-3) & 5.02e-1 (1.43e-3) & 9.31e-3 (2.58e-3) & 1.25e+1 (2.30e+0) & 7.76e-3 (2.46e-3) & 7.07e-2 (2.11e-2) \\
& H=256  & 2.46e-2 (5.52e-3) & 5.02e-1 (1.43e-3) & 9.81e-3 (2.12e-3) & 1.05e+1 (4.44e+0) & 6.17e-3 (3.04e-3) & 6.26e-2 (7.76e-3) \\
& H=512  & 1.74e-2 (2.85e-3) & 5.02e-1 (1.43e-3) & 3.42e-3 (1.19e-3) & 9.00e+0 (3.43e+0) & 7.63e-3 (3.33e-3) & 4.69e-2 (8.57e-3) \\
& H=1024 & 1.14e-2 (2.01e-3) & 5.02e-1 (1.43e-3) & 3.02e-3 (5.51e-4) & 1.48e+0 (1.06e+0) & 2.15e-2 (5.78e-3) & 9.22e-2 (3.42e-2) \\
\hline

GP-HM \cite{fang2024gaussian} & Grid=100$^2$ & 6.72e-1 (3.27e-3) & \textbf{2.22e-1 (1.41e-3)} & 8.67e-1 (1.25e-3) & 4.66e-1 (1.10e-3) & 4.99e-1 (2.71e-3) & \textbf{9.45e-5 (6.28e-7)} \\
\hline

\multirow{4}{*}{ConFIG \cite{ConFIG2025}}
& H=128  & 6.88e-4 (2.14e-4) & 5.02e-1 (1.43e-3) & 1.57e-3 (1.84e-3) & \textbf{1.01e-1 (1.04e-1)} & 1.84e-2 (5.12e-3) & 3.70e-2 (2.64e-3) \\
& H=256  & 1.82e-3 (1.11e-3) & 5.02e-1 (1.44e-3) & 4.05e-4 (3.74e-4) & 4.16e-1 (4.62e-1) & 1.68e-2 (8.26e-3) & 2.74e-2 (5.19e-3) \\
& H=512  & 7.76e-4 (5.33e-4) & 5.02e-1 (1.43e-3) & 2.08e-3 (1.92e-3) & 2.96e-1 (2.09e-1) & 2.00e-2 (9.94e-3) & 1.74e-2 (3.75e-3) \\
& H=1024 & 7.68e-4 (3.78e-4) & 5.02e-1 (1.43e-3) & 9.38e-3 (9.19e-3) & 1.39e-1 (1.05e-1) & 1.79e-2 (7.67e-3) & 2.25e-2 (2.48e-3) \\
\hline

\multirow{4}{*}{PINN DCGD \cite{DCGD2024}}
& H=128  & 2.20e-3 (9.64e-4) & 5.02e-1 (1.44e-3) & 5.38e-4 (4.56e-4) & 1.58e-1 (1.92e-1) & 2.04e-2 (5.05e-3) & 3.36e-2 (3.28e-3) \\
& H=256  & 6.39e-3 (7.17e-3) & 5.02e-1 (1.44e-3) & 1.74e-3 (8.38e-4) & 4.63e-1 (3.64e-1) & 1.90e-2 (4.99e-3) & 2.10e-2 (2.33e-3) \\
& H=512  & 3.06e-3 (2.15e-3) & 5.02e-1 (1.44e-3) & 3.79e-3 (3.49e-3) & 8.31e-1 (4.01e-1) & 2.82e-2 (9.29e-3) & 1.67e-2 (2.50e-3) \\
& H=1024 & 1.16e-2 (5.86e-3) & 5.02e-1 (1.43e-3) & 1.05e-2 (8.97e-3) & 1.93e-1 (1.77e-1) & 2.20e-2 (6.21e-3) & 4.39e-2 (2.71e-2) \\
\hline
\end{tabularx}
\end{table}

\section{Discussion}
\label{sec:5}
We now look at the limits of MetaColloc. We focus on the extremely high-frequency Helmholtz equation. MetaColloc solves smooth and non-linear equations well. It also handles high-order derivatives in these settings. However, it does not reach the same level at the highest frequencies. We first thought the linear solver caused this problem. We suspected that the high-frequency branch created large condition numbers. We tested this idea. We found that the condition number stays near $10^9$ for both good and bad solves. The condition number does not explain the failure. We must look deeper.

\begin{figure}[ht]
    \centering
    \includegraphics[width=\textwidth]{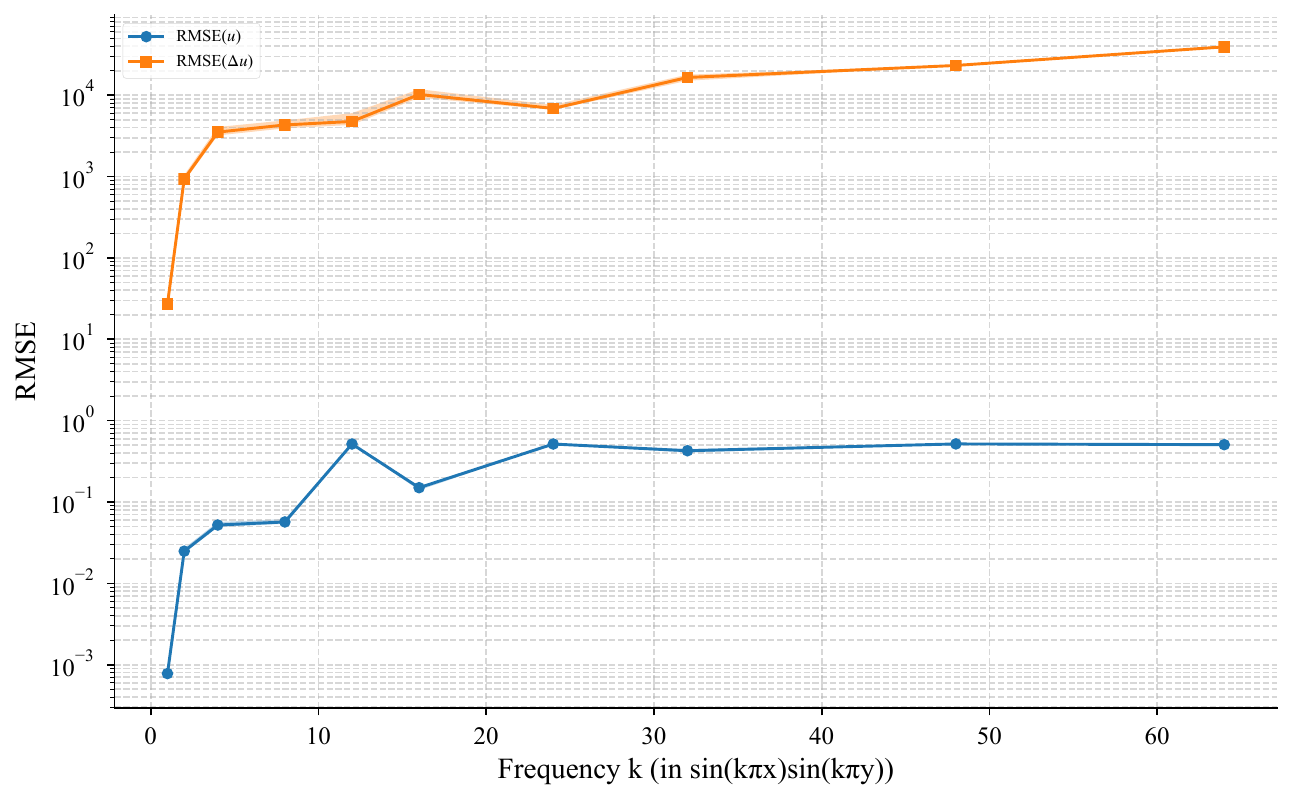}
    \caption{\textbf{Frequency sweep exposes the spectral limitations of MetaColloc.} Although the meta-learned basis can represent high-frequency modes in function space, its Laplacian responses become increasingly unstable as frequency grows. The divergence between $RMSE(u)$ and $RMSE(\Delta u)$ highlights a structural gap between interpolation accuracy and PDE operator compatibility. This phenomenon reflects a general property of function-only meta-objectives rather than a failure of the architecture, and suggests a natural direction for operator-aware meta-learning.}
    \label{fig:frequency}
\end{figure}

We design a frequency sweep to find the real cause. Figure~\ref{fig:frequency} shows the results. We measure the error of the function, in which H is fixed at 512. We also measure the error of its Laplacian. The gap is large. The basis fits the function values well, even at high frequencies. However, the Laplacian error grows quickly. It becomes three to five orders of magnitude larger. The basis matches the target points, but it moves sharply between them. The PDE operator amplifies these small changes. This creates large errors in the final solution.

This behavior comes from our training setup. We train on Gaussian Random Fields. We only ask the model to match function values. We do not ask it to keep its derivatives stable. The model learns to fit many shapes. But it does not learn how these shapes behave under differential operators. This creates a gap between function approximation and operator stability. We call this an ``operator–function mismatch''. The linear solver then works in a space where small changes in the basis create large changes in the PDE residual. The solver cannot fix this mismatch.

This limitation gives us a clear next step. We need operator-aware meta-training. We need to guide the basis so that it stays stable under the PDE operators we care about. We can do this during the offline stage. We do not need to change the test-time pipeline. We do not need to train a new network for each PDE. We do not move toward a standard PINN. We keep the optimization-free nature of MetaColloc. We keep the single-step solve. We only improve the basis so that it works better at extremely high frequencies.

MetaColloc already solves many PDEs quickly and effectively. It also provides a new way to think about PDE solvers. The frequency sweep reveals a deeper structure behind these problems. It shows that function fitting and operator stability are not the same. This insight opens a new direction. We can now design basis functions that perform well in both spaces. We believe this will lead to stronger and more stable zero-shot PDE solvers.

\section{Conclusion}
\label{sec:6}
We presented MetaColloc, a new framework for solving PDEs that removes the need to train a separate network for each equation. By decoupling basis discovery from solution, we first learn a universal neural basis dictionary offline using a dual-branch network trained on multi-scale Gaussian Random Fields, without any PDE training data. This frozen dictionary is then used online to solve new PDEs.

For linear PDEs, MetaColloc reduces inference to a single linear least-squares solve, completely avoiding gradient descent. For nonlinear PDEs, we use Newton-Raphson iterations to obtain fast and stable quadratic convergence. Experiments show large gains in both accuracy and speed on smooth and nonlinear equations, and the method scales well to three-dimensional problems.

We also identify a clear limitation: MetaColloc struggles on extremely high-frequency physics. Our frequency-sweep analysis shows that matching function values alone is not sufficient; the learned basis must also remain stable under differential operators. This points to operator-aware meta-learning as a key direction for future work. Overall, MetaColloc turns a slow optimization problem into fast linear algebra, providing a strong optimization-free and data-free foundation for future PDE solvers.

\bibliography{neurips_2026}

\begin{thebibliography}{50}
\providecommand{\natexlab}[1]{#1}
\providecommand{\url}[1]{\texttt{#1}}
\expandafter\ifx\csname urlstyle\endcsname\relax
  \providecommand{\doi}[1]{doi: #1}\else
  \providecommand{\doi}{doi: \begingroup \urlstyle{rm}\Url}\fi

\bibitem[Alkin et~al.(2024)Alkin, F\"{u}rst, Schmid, Gruber, Holzleitner, and Brandstetter]{UPT2024}
Benedikt Alkin, Andreas F\"{u}rst, Simon Schmid, Lukas Gruber, Markus Holzleitner, and Johannes Brandstetter.
\newblock Universal physics transformers: A framework for efficiently scaling neural operators.
\newblock In A.~Globerson, L.~Mackey, D.~Belgrave, A.~Fan, U.~Paquet, J.~Tomczak, and C.~Zhang, editors, \emph{Advances in Neural Information Processing Systems}, volume~37, pages 25152--25194. Curran Associates, Inc., 2024.
\newblock \doi{10.52202/079017-0793}.
\newblock URL \url{https://proceedings.neurips.cc/paper_files/paper/2024/file/2cd36d327f33d47b372d4711edd08de0-Paper-Conference.pdf}.

\bibitem[Aronszajn(1950)]{aronszajn1950theory}
Nachman Aronszajn.
\newblock Theory of reproducing kernels.
\newblock \emph{Transactions of the American mathematical society}, 68\penalty0 (3):\penalty0 337--404, 1950.

\bibitem[Bastek et~al.(2025)Bastek, Sun, and Kochmann]{PhysicsInformedDiffusion2025}
Jan-Hendrik Bastek, WaiChing Sun, and Dennis Kochmann.
\newblock Physics-informed diffusion models.
\newblock In \emph{The Thirteenth International Conference on Learning Representations}, 2025.
\newblock URL \url{https://openreview.net/forum?id=tpYeermigp}.

\bibitem[Bhatia et~al.(2025)Bhatia, Koehler, and Thuerey]{PRDP2025}
Kanishk Bhatia, Felix Koehler, and Nils Thuerey.
\newblock {PRDP}: Progressively refined differentiable physics.
\newblock In \emph{The Thirteenth International Conference on Learning Representations}, 2025.
\newblock URL \url{https://openreview.net/forum?id=9Fh0z1JmPU}.

\bibitem[Chen et~al.(2021)]{chen2021solving}
Yuntian Chen et~al.
\newblock Solving partial differential equations with point source based on physics-informed neural networks.
\newblock \emph{Journal of Computational Physics}, 2021.

\bibitem[Chowdhery et~al.(2022)Chowdhery, Narang, Devlin, Bosma, Mishra, Roberts, Barham, Chung, Sutton, Gehrmann, Schuh, Shi, Tsvyashchenko, Maynez, Rao, Barnes, Tay, Shazeer, Prabhakaran, Reif, Du, Hutchinson, Pope, Bradbury, Austin, Isard, Gur-Ari, Yin, Duke, Levskaya, Ghemawat, Dev, Michalewski, Garcia, Misra, Robinson, Fedus, Zhou, Ippolito, Luan, Lim, Zoph, Spiridonov, Sepassi, Dohan, Agrawal, Omernick, Dai, Pillai, Pellat, Lewkowycz, Moreira, Child, Polozov, Lee, Zhou, Wang, Saeta, Diaz, Firat, Catasta, Wei, Meier-Hellstern, Eck, Dean, Petrov, and Fiedel]{chowdhery2022palmscalinglanguagemodeling}
Aakanksha Chowdhery, Sharan Narang, Jacob Devlin, Maarten Bosma, Gaurav Mishra, Adam Roberts, Paul Barham, Hyung~Won Chung, Charles Sutton, Sebastian Gehrmann, Parker Schuh, Kensen Shi, Sasha Tsvyashchenko, Joshua Maynez, Abhishek Rao, Parker Barnes, Yi~Tay, Noam Shazeer, Vinodkumar Prabhakaran, Emily Reif, Nan Du, Ben Hutchinson, Reiner Pope, James Bradbury, Jacob Austin, Michael Isard, Guy Gur-Ari, Pengcheng Yin, Toju Duke, Anselm Levskaya, Sanjay Ghemawat, Sunipa Dev, Henryk Michalewski, Xavier Garcia, Vedant Misra, Kevin Robinson, Liam Fedus, Denny Zhou, Daphne Ippolito, David Luan, Hyeontaek Lim, Barret Zoph, Alexander Spiridonov, Ryan Sepassi, David Dohan, Shivani Agrawal, Mark Omernick, Andrew~M. Dai, Thanumalayan~Sankaranarayana Pillai, Marie Pellat, Aitor Lewkowycz, Erica Moreira, Rewon Child, Oleksandr Polozov, Katherine Lee, Zongwei Zhou, Xuezhi Wang, Brennan Saeta, Mark Diaz, Orhan Firat, Michele Catasta, Jason Wei, Kathy Meier-Hellstern, Douglas Eck, Jeff Dean, Slav Petrov, and Noah Fiedel.
\newblock Palm: Scaling language modeling with pathways, 2022.
\newblock URL \url{https://arxiv.org/abs/2204.02311}.

\bibitem[Dangel et~al.(2024)Dangel, M\"{u}ller, and Zeinhofer]{KFAC2024}
Felix Dangel, Johannes M\"{u}ller, and Marius Zeinhofer.
\newblock Kronecker-factored approximate curvature for physics-informed neural networks.
\newblock In A.~Globerson, L.~Mackey, D.~Belgrave, A.~Fan, U.~Paquet, J.~Tomczak, and C.~Zhang, editors, \emph{Advances in Neural Information Processing Systems}, volume~37, pages 34582--34636. Curran Associates, Inc., 2024.
\newblock \doi{10.52202/079017-1091}.
\newblock URL \url{https://proceedings.neurips.cc/paper_files/paper/2024/file/3d27d607586984908900eaa8ce19c96c-Paper-Conference.pdf}.

\bibitem[Dong and Li(2021)]{dong2021local}
Suchuan Dong and Zongyi Li.
\newblock Local extreme learning machines and domain decomposition for solving linear and nonlinear partial differential equations.
\newblock \emph{Computer Methods in Applied Mechanics and Engineering}, 387:\penalty0 114129, 2021.
\newblock \doi{10.1016/j.cma.2021.114129}.

\bibitem[Fang et~al.(2024)Fang, Cooley, Long, Li, Kirby, and Zhe]{fang2024gaussian}
Shikai Fang, Madison Cooley, Da~Long, Shibo Li, Mike Kirby, and Shandian Zhe.
\newblock Solving high frequency and multi-scale {PDE}s with gaussian processes.
\newblock In \emph{The Twelfth International Conference on Learning Representations}, 2024.
\newblock URL \url{https://openreview.net/forum?id=q4AEBLHuA6}.

\bibitem[Feng et~al.(2025)Feng, Huang, Liao, Liu, Liu, and Yan]{SINGER2025}
Mingquan Feng, Yixin Huang, Weixin Liao, Yuhong Liu, Yizhou Liu, and Junchi Yan.
\newblock {SINGER}: Stochastic network graph evolving operator for high dimensional {PDE}s.
\newblock In \emph{The Thirteenth International Conference on Learning Representations}, 2025.
\newblock URL \url{https://openreview.net/forum?id=wVADj7yKee}.

\bibitem[Gattiglio et~al.(2024)Gattiglio, Grigoryeva, and Tamborrino]{RandNetParareal2024}
Guglielmo Gattiglio, Lyudmila Grigoryeva, and Massimiliano Tamborrino.
\newblock Randnet-parareal: a time-parallel pde solver using random neural networks.
\newblock In A.~Globerson, L.~Mackey, D.~Belgrave, A.~Fan, U.~Paquet, J.~Tomczak, and C.~Zhang, editors, \emph{Advances in Neural Information Processing Systems}, volume~37, pages 94993--95025. Curran Associates, Inc., 2024.
\newblock \doi{10.52202/079017-3011}.
\newblock URL \url{https://proceedings.neurips.cc/paper_files/paper/2024/file/acb94e709f02895fd98b5867f0b184f3-Paper-Conference.pdf}.

\bibitem[Gavrikov and Keuper(2023)]{gavrikov2023powerlinearcombinationslearning}
Paul Gavrikov and Janis Keuper.
\newblock The power of linear combinations: Learning with random convolutions, 2023.
\newblock URL \url{https://arxiv.org/abs/2301.11360}.

\bibitem[Ha~Quang(2006)]{ha2006reproducing}
Minh Ha~Quang.
\newblock \emph{Reproducing kernel Hilbert spaces in learning theory}.
\newblock Brown University, 2006.

\bibitem[Hao et~al.(2024)Hao, Liu, and Yang]{NINO2024}
Wenrui Hao, Xinliang Liu, and Yahong Yang.
\newblock Newton informed neural operator for solving nonlinear partial differential equations.
\newblock In A.~Globerson, L.~Mackey, D.~Belgrave, A.~Fan, U.~Paquet, J.~Tomczak, and C.~Zhang, editors, \emph{Advances in Neural Information Processing Systems}, volume~37, pages 120832--120860. Curran Associates, Inc., 2024.
\newblock \doi{10.52202/079017-3839}.
\newblock URL \url{https://proceedings.neurips.cc/paper_files/paper/2024/file/dae8afc6b990aa0b3b5efaa096fbd7fa-Paper-Conference.pdf}.

\bibitem[Henderson(2024)]{henderson2024sobolev}
Iain Henderson.
\newblock Sobolev regularity of gaussian random fields.
\newblock \emph{Journal of Functional Analysis}, 286\penalty0 (3):\penalty0 110241, 2024.

\bibitem[Herde et~al.(2024)Herde, Raoni\'{c}, Rohner, K\"{a}ppeli, Molinaro, de~B\'{e}zenac, and Mishra]{Poseidon2024}
Maximilian Herde, Bogdan Raoni\'{c}, Tobias Rohner, Roger K\"{a}ppeli, Roberto Molinaro, Emmanuel de~B\'{e}zenac, and Siddhartha Mishra.
\newblock Poseidon: Efficient foundation models for pdes.
\newblock In A.~Globerson, L.~Mackey, D.~Belgrave, A.~Fan, U.~Paquet, J.~Tomczak, and C.~Zhang, editors, \emph{Advances in Neural Information Processing Systems}, volume~37, pages 72525--72624. Curran Associates, Inc., 2024.
\newblock \doi{10.52202/079017-2311}.
\newblock URL \url{https://proceedings.neurips.cc/paper_files/paper/2024/file/84e1b1ec17bb11c57234e96433022a9a-Paper-Conference.pdf}.

\bibitem[Huang et~al.(2006)Huang, Zhu, and Siew]{huang2006extreme}
Guang-Bin Huang, Qin-Yu Zhu, and Chee-Kheong Siew.
\newblock Extreme learning machine: theory and applications.
\newblock In \emph{Neurocomputing}, 2006.

\bibitem[Huang et~al.(2024)Huang, Yang, Wang, and Park]{DiffusionPDE2024}
Jiahe Huang, Guandao Yang, Zichen Wang, and Jeong~Joon Park.
\newblock Diffusionpde: Generative pde-solving under partial observation.
\newblock In A.~Globerson, L.~Mackey, D.~Belgrave, A.~Fan, U.~Paquet, J.~Tomczak, and C.~Zhang, editors, \emph{Advances in Neural Information Processing Systems}, volume~37, pages 130291--130323. Curran Associates, Inc., 2024.
\newblock \doi{10.52202/079017-4140}.
\newblock URL \url{https://proceedings.neurips.cc/paper_files/paper/2024/file/eb3878c1dcbfff9ee95d5d033e5f5942-Paper-Conference.pdf}.

\bibitem[Hwang and Lim(2024)]{DCGD2024}
Youngsik Hwang and Dong-Young Lim.
\newblock Dual cone gradient descent for training physics-informed neural networks.
\newblock In A.~Globerson, L.~Mackey, D.~Belgrave, A.~Fan, U.~Paquet, J.~Tomczak, and C.~Zhang, editors, \emph{Advances in Neural Information Processing Systems}, volume~37, pages 98563--98595. Curran Associates, Inc., 2024.
\newblock \doi{10.52202/079017-3128}.
\newblock URL \url{https://proceedings.neurips.cc/paper_files/paper/2024/file/b2b781badeeb49896c4b324c466ec442-Paper-Conference.pdf}.

\bibitem[Kang et~al.(2025)Kang, Oh, Hong, and Park]{PIG2025}
Namgyu Kang, Jaemin Oh, Youngjoon Hong, and Eunbyung Park.
\newblock {PIG}: Physics-informed gaussians as adaptive parametric mesh representations.
\newblock In \emph{The Thirteenth International Conference on Learning Representations}, 2025.
\newblock URL \url{https://openreview.net/forum?id=y5B0ca4mjt}.

\bibitem[Li et~al.(2021)Li, Kovachki, Azizzadenesheli, liu, Bhattacharya, Stuart, and Anandkumar]{li2020fourier}
Zongyi Li, Nikola~Borislavov Kovachki, Kamyar Azizzadenesheli, Burigede liu, Kaushik Bhattacharya, Andrew Stuart, and Anima Anandkumar.
\newblock Fourier neural operator for parametric partial differential equations.
\newblock In \emph{International Conference on Learning Representations}, 2021.
\newblock URL \url{https://openreview.net/forum?id=c8P9NQVtmnO}.

\bibitem[Li et~al.(2023)Li, Huang, Liu, and Anandkumar]{GeoFNO2024}
Zongyi Li, Daniel~Zhengyu Huang, Burigede Liu, and Anima Anandkumar.
\newblock Fourier neural operator with learned deformations for pdes on general geometries.
\newblock \emph{Journal of Machine Learning Research}, 24\penalty0 (388):\penalty0 1--26, 2023.
\newblock URL \url{http://jmlr.org/papers/v24/23-0064.html}.

\bibitem[Liao(2025)]{liao2025solving}
Chunyang Liao.
\newblock Solving partial differential equations with random feature models.
\newblock \emph{Communications in Nonlinear Science and Numerical Simulation}, page 109343, 2025.

\bibitem[Liu et~al.(2025{\natexlab{a}})Liu, Chu, and Thuerey]{ConFIG2025}
Qiang Liu, Mengyu Chu, and Nils Thuerey.
\newblock Con{FIG}: Towards conflict-free training of physics informed neural networks.
\newblock In \emph{The Thirteenth International Conference on Learning Representations}, 2025{\natexlab{a}}.
\newblock URL \url{https://openreview.net/forum?id=APojAzJQiq}.

\bibitem[Liu et~al.(2025{\natexlab{b}})Liu, Wang, Vaidya, Ruehle, Halverson, Soljacic, Hou, and Tegmark]{KAN2025}
Ziming Liu, Yixuan Wang, Sachin Vaidya, Fabian Ruehle, James Halverson, Marin Soljacic, Thomas~Y. Hou, and Max Tegmark.
\newblock {KAN}: Kolmogorov{\textendash}arnold networks.
\newblock In \emph{The Thirteenth International Conference on Learning Representations}, 2025{\natexlab{b}}.
\newblock URL \url{https://openreview.net/forum?id=Ozo7qJ5vZi}.

\bibitem[Loshchilov and Hutter(2019)]{loshchilov2018decoupled}
Ilya Loshchilov and Frank Hutter.
\newblock Decoupled weight decay regularization.
\newblock In \emph{International Conference on Learning Representations}, 2019.
\newblock URL \url{https://openreview.net/forum?id=Bkg6RiCqY7}.

\bibitem[Lu et~al.(2021)Lu, Jin, Pang, Zhang, and Karniadakis]{lu2021learning}
Lu~Lu, Pengzhan Jin, Guofei Pang, Zhongqiang Zhang, and George~Em Karniadakis.
\newblock Learning nonlinear operators via deeponet based on the universal approximation theorem of operators.
\newblock \emph{Nature machine intelligence}, 3\penalty0 (3):\penalty0 218--229, 2021.
\newblock \doi{10.1038/s42256-021-00302-5}.

\bibitem[Lu et~al.(2025)Lu, Ju, and Zhu]{LU2025113902}
Tianzheng Lu, Lili Ju, and Liyong Zhu.
\newblock A multiple transferable neural network method with domain decomposition for elliptic interface problems.
\newblock \emph{Journal of Computational Physics}, 530:\penalty0 113902, 2025.
\newblock ISSN 0021-9991.
\newblock \doi{https://doi.org/10.1016/j.jcp.2025.113902}.
\newblock URL \url{https://www.sciencedirect.com/science/article/pii/S0021999125001858}.

\bibitem[Luki{\'c} and Beder(2001)]{lukic2001stochastic}
Milan Luki{\'c} and Jay Beder.
\newblock Stochastic processes with sample paths in reproducing kernel hilbert spaces.
\newblock \emph{Transactions of the American Mathematical Society}, 353\penalty0 (10):\penalty0 3945--3969, 2001.

\bibitem[McCabe et~al.(2024)McCabe, R\'{e}galdo-Saint~Blancard, Parker, Ohana, Cranmer, Bietti, Eickenberg, Golkar, Krawezik, Lanusse, Pettee, Tesileanu, Cho, and Ho]{MPP2024}
Michael McCabe, Bruno R\'{e}galdo-Saint~Blancard, Liam Parker, Ruben Ohana, Miles Cranmer, Alberto Bietti, Michael Eickenberg, Siavash Golkar, Geraud Krawezik, Francois Lanusse, Mariel Pettee, Tiberiu Tesileanu, Kyunghyun Cho, and Shirley Ho.
\newblock Multiple physics pretraining for spatiotemporal surrogate models.
\newblock In A.~Globerson, L.~Mackey, D.~Belgrave, A.~Fan, U.~Paquet, J.~Tomczak, and C.~Zhang, editors, \emph{Advances in Neural Information Processing Systems}, volume~37, pages 119301--119335. Curran Associates, Inc., 2024.
\newblock \doi{10.52202/079017-3791}.
\newblock URL \url{https://proceedings.neurips.cc/paper_files/paper/2024/file/d7cb9db5ade2db7814fbd01ee59f4c7b-Paper-Conference.pdf}.

\bibitem[Rahaman et~al.(2019)Rahaman, Baratin, Arpit, Eberhardt, Yoshua, and Courville]{rahaman2019spectral}
Nasim Rahaman, Aristide Baratin, Devansh Arpit, Felix Eberhardt, Bengio Yoshua, and Aaron Courville.
\newblock On the spectral bias of neural networks.
\newblock In \emph{International Conference on Machine Learning}, pages 5301--5310. PMLR, 2019.

\bibitem[Rahimi and Recht(2007)]{rahimi2007random}
Ali Rahimi and Benjamin Recht.
\newblock Random features for large-scale kernel machines.
\newblock In \emph{Advances in neural information processing systems}, 2007.

\bibitem[Rahman et~al.(2024)Rahman, George, Elleithy, Leibovici, Li, Bonev, White, Berner, Yeh, Kossaifi, Azizzadenesheli, and Anandkumar]{CoDANO2024}
Ashiqur Rahman, Robert~Joseph George, Mogab Elleithy, Daniel Leibovici, Zongyi Li, Boris Bonev, Colin White, Julius Berner, Raymond~A. Yeh, Jean Kossaifi, Kamyar Azizzadenesheli, and Anima Anandkumar.
\newblock Pretraining codomain attention neural operators for solving multiphysics pdes.
\newblock In A.~Globerson, L.~Mackey, D.~Belgrave, A.~Fan, U.~Paquet, J.~Tomczak, and C.~Zhang, editors, \emph{Advances in Neural Information Processing Systems}, volume~37, pages 104035--104064. Curran Associates, Inc., 2024.
\newblock \doi{10.52202/079017-3306}.
\newblock URL \url{https://proceedings.neurips.cc/paper_files/paper/2024/file/bc75fa9843a7905bbed9d83895a88f7f-Paper-Conference.pdf}.

\bibitem[Raissi et~al.(2019)Raissi, Perdikaris, and Karniadakis]{raissi2019physics}
M.~Raissi, P.~Perdikaris, and G.E. Karniadakis.
\newblock Physics-informed neural networks: A deep learning framework for solving forward and inverse problems involving nonlinear partial differential equations.
\newblock \emph{Journal of Computational Physics}, 378:\penalty0 686--707, 2019.
\newblock ISSN 0021-9991.
\newblock \doi{https://doi.org/10.1016/j.jcp.2018.10.045}.
\newblock URL \url{https://www.sciencedirect.com/science/article/pii/S0021999118307125}.

\bibitem[Serrano et~al.(2024)Serrano, Wang, Le~Naour, Vittaut, and Gallinari]{AROMA2024}
Louis Serrano, Thomas~X Wang, Etienne Le~Naour, Jean-No\"{e}l Vittaut, and Patrick Gallinari.
\newblock Aroma: Preserving spatial structure for latent pde modeling with local neural fields.
\newblock In A.~Globerson, L.~Mackey, D.~Belgrave, A.~Fan, U.~Paquet, J.~Tomczak, and C.~Zhang, editors, \emph{Advances in Neural Information Processing Systems}, volume~37, pages 13489--13521. Curran Associates, Inc., 2024.
\newblock \doi{10.52202/079017-0431}.
\newblock URL \url{https://proceedings.neurips.cc/paper_files/paper/2024/file/185a120a3f709187e68bd092e6098851-Paper-Conference.pdf}.

\bibitem[Shehata et~al.(2025)Shehata, Holzschuh, and Thuerey]{TSM2025}
Youssef Shehata, Benjamin Holzschuh, and Nils Thuerey.
\newblock Improved sampling of diffusion models in fluid dynamics with tweedie's formula.
\newblock In \emph{The Thirteenth International Conference on Learning Representations}, 2025.
\newblock URL \url{https://openreview.net/forum?id=0FbzC7B9xI}.

\bibitem[Shysheya et~al.(2024)Shysheya, Diaconu, Bergamin, Perdikaris, Hern\'{a}ndez-Lobato, Turner, and Mathieu]{ConditionalDiffusion2024}
Aliaksandra Shysheya, Cristiana Diaconu, Federico Bergamin, Paris Perdikaris, Jos\'{e}~Miguel Hern\'{a}ndez-Lobato, Richard~E. Turner, and Emile Mathieu.
\newblock On conditional diffusion models for pde simulations.
\newblock In A.~Globerson, L.~Mackey, D.~Belgrave, A.~Fan, U.~Paquet, J.~Tomczak, and C.~Zhang, editors, \emph{Advances in Neural Information Processing Systems}, volume~37, pages 23246--23300. Curran Associates, Inc., 2024.
\newblock \doi{10.52202/079017-0732}.
\newblock URL \url{https://proceedings.neurips.cc/paper_files/paper/2024/file/2974844555dc383ea16c5f35833c7a57-Paper-Conference.pdf}.

\bibitem[Tancik et~al.(2020)Tancik, Srinivasan, Mildenhall, Fridovich-Keil, Raghavan, Singhal, Ramamoorthi, Barron, and Ng]{tancik2020fourier}
Matthew Tancik, Pratul Srinivasan, Ben Mildenhall, Sara Fridovich-Keil, Nithin Raghavan, Utkarsh Singhal, Ravi Ramamoorthi, Jonathan Barron, and Ren Ng.
\newblock Fourier features let networks learn high frequency functions in low dimensional domains.
\newblock In H.~Larochelle, M.~Ranzato, R.~Hadsell, M.F. Balcan, and H.~Lin, editors, \emph{Advances in Neural Information Processing Systems}, volume~33, pages 7537--7547. Curran Associates, Inc., 2020.
\newblock URL \url{https://proceedings.neurips.cc/paper_files/paper/2020/file/55053683268957697aa39fba6f231c68-Paper.pdf}.

\bibitem[Wandel et~al.(2025)Wandel, Schulz, and Klein]{Metamizer2025}
Nils Wandel, Stefan Schulz, and Reinhard Klein.
\newblock Metamizer: A versatile neural optimizer for fast and accurate physics simulations.
\newblock In \emph{The Thirteenth International Conference on Learning Representations}, 2025.
\newblock URL \url{https://openreview.net/forum?id=60TXv9Xif5}.

\bibitem[Wang et~al.(2021{\natexlab{a}})Wang, Teng, and Perdikaris]{wang2020understanding}
Sifan Wang, Yujun Teng, and Paris Perdikaris.
\newblock Understanding and mitigating gradient flow pathologies in physics-informed neural networks.
\newblock In \emph{SIAM Journal on Scientific Computing}, 2021{\natexlab{a}}.

\bibitem[Wang et~al.(2021{\natexlab{b}})Wang, Wang, and Perdikaris]{wang2021eigenvector}
Sifan Wang, Hanwen Wang, and Paris Perdikaris.
\newblock On the eigenvector bias of fourier feature networks: From regression to solving multi-scale pdes with physics-informed neural networks.
\newblock In \emph{Computer Methods in Applied Mechanics and Engineering}, 2021{\natexlab{b}}.

\bibitem[Wang et~al.(2022)Wang, Sankaran, and Perdikaris]{wang2022respecting}
Sifan Wang, Shyam Sankaran, and Paris Perdikaris.
\newblock Respecting causality is all you need for training physics-informed neural networks.
\newblock In \emph{arXiv preprint arXiv:2203.07404}, 2022.

\bibitem[Wang et~al.(2025)Wang, Seidman, Sankaran, Wang, Pappas, and Perdikaris]{CViT2025}
Sifan Wang, Jacob~H Seidman, Shyam Sankaran, Hanwen Wang, George~J. Pappas, and Paris Perdikaris.
\newblock {CV}it: Continuous vision transformer for operator learning.
\newblock In \emph{The Thirteenth International Conference on Learning Representations}, 2025.
\newblock URL \url{https://openreview.net/forum?id=cRnCcuLvyr}.

\bibitem[Wang and Wang(2024)]{LNO2024}
Tian Wang and Chuang Wang.
\newblock Latent neural operator for solving forward and inverse pde problems.
\newblock In A.~Globerson, L.~Mackey, D.~Belgrave, A.~Fan, U.~Paquet, J.~Tomczak, and C.~Zhang, editors, \emph{Advances in Neural Information Processing Systems}, volume~37, pages 33085--33107. Curran Associates, Inc., 2024.
\newblock \doi{10.52202/079017-1042}.
\newblock URL \url{https://proceedings.neurips.cc/paper_files/paper/2024/file/39f6d5c2e310a5a629dcfc4d517aa0d1-Paper-Conference.pdf}.

\bibitem[Wei et~al.(2025)Wei, Feng, Yang, Feng, Hu, Zheng, Zhang, Fan, and Wu]{CLDiffPhyCon2025}
Long Wei, Haodong Feng, Yuchen Yang, Ruiqi Feng, Peiyan Hu, Xiang Zheng, Tao Zhang, Dixia Fan, and Tailin Wu.
\newblock {CL}-diffphycon: Closed-loop diffusion control of complex physical systems.
\newblock In \emph{The Thirteenth International Conference on Learning Representations}, 2025.
\newblock URL \url{https://openreview.net/forum?id=PiHGrTTnvb}.

\bibitem[Williams and Rasmussen(2006)]{williams2006gaussian}
Christopher~KI Williams and Carl~Edward Rasmussen.
\newblock \emph{Gaussian processes for machine learning}, volume~2.
\newblock MIT press Cambridge, MA, 2006.

\bibitem[Wu et~al.(2024)Wu, Luo, Ma, Wang, and Long]{RoPINN2024}
Haixu Wu, Huakun Luo, Yuezhou Ma, Jianmin Wang, and Mingsheng Long.
\newblock Ropinn: Region optimized physics-informed neural networks.
\newblock In A.~Globerson, L.~Mackey, D.~Belgrave, A.~Fan, U.~Paquet, J.~Tomczak, and C.~Zhang, editors, \emph{Advances in Neural Information Processing Systems}, volume~37, pages 110494--110532. Curran Associates, Inc., 2024.
\newblock \doi{10.52202/079017-3508}.
\newblock URL \url{https://proceedings.neurips.cc/paper_files/paper/2024/file/c745bfa5b50544882938ff4f89ff26ac-Paper-Conference.pdf}.

\bibitem[Zhang et~al.(2024)Zhang, Bao, Ju, and Zhang]{zhang2024transferable}
Zezhong Zhang, Feng Bao, Lili Ju, and Guannan Zhang.
\newblock Transferable neural networks for partial differential equations.
\newblock \emph{Journal of Scientific Computing}, 99\penalty0 (2):\penalty0 31, 2024.
\newblock \doi{10.1007/s10915-024-02463-y}.

\bibitem[Zheng et~al.(2024)Zheng, Li, Xu, Zhu, Lin, and Zhang]{MambaNO2024}
Jianwei Zheng, Wei Li, Ni~Xu, Junwei Zhu, Xiaoxu Lin, and Xiaoqin Zhang.
\newblock Alias-free mamba neural operator.
\newblock In A.~Globerson, L.~Mackey, D.~Belgrave, A.~Fan, U.~Paquet, J.~Tomczak, and C.~Zhang, editors, \emph{Advances in Neural Information Processing Systems}, volume~37, pages 52962--52995. Curran Associates, Inc., 2024.
\newblock \doi{10.52202/079017-1678}.
\newblock URL \url{https://proceedings.neurips.cc/paper_files/paper/2024/file/5ee553ec47c31e46a1209bb858b30aa5-Paper-Conference.pdf}.

\bibitem[Zhou et~al.(2025)Zhou, Li, Schneier, Jr, and Farimani]{Text2PDE2025}
Anthony Zhou, Zijie Li, Michael Schneier, John R~Buchanan Jr, and Amir~Barati Farimani.
\newblock Text2{PDE}: Latent diffusion models for accessible physics simulation.
\newblock In \emph{The Thirteenth International Conference on Learning Representations}, 2025.
\newblock URL \url{https://openreview.net/forum?id=Nb3a8aUGfj}.

\end{thebibliography}
\bibliographystyle{plainnat}

\clearpage
\appendix

\section{Experimental Setup and Hyperparameters}
\label{app:a}
We provide the exact settings used to train and evaluate MetaColloc. We designed these parameters to ensure full reproducibility.

We use double precision (FP64) for all matrix solving steps. This high precision is crucial. It prevents condition number explosion during the linear least-squares inversions. We run all meta-training and evaluation scripts on a single NVIDIA GPU. The python version is 3.12.12. The PyTorch version is 2.6.0+cu124. The GPU is NVIDIA A800-SXM4-80GB.

During the offline meta-training phase, we sample functions from a multi-scale Gaussian Random Field distribution. We generate tasks by mixing three probability modes. We use a pure Radial Basis Function mode 40\% of the time. We use a high-frequency mode 40\% of the time. We mix both modes for the remaining 20\% of the tasks. For smooth functions, we draw length scales randomly between 0.005 and 0.05. For high-frequency functions, we sample center frequencies uniformly between 10.0 and 300.0. We add a noise bandwidth between 1.0 and 15.0 to create complex signal aliases. 

We train the neural network for 1000 epochs. Each epoch contains 128 random GRF tasks. For each task, we sample 4000 training points and 1500 testing points uniformly across the domain. We use the AdamW \cite{loshchilov2018decoupled} optimizer. We set the initial learning rate to $10^{-3}$ and apply a weight decay of $10^{-4}$. We use a cosine annealing scheduler to reduce the learning rate gradually. 

During the online test phase, we evaluate the solver on the specific PDEs. We sample 2000 points randomly inside the spatial domain. We sample 300 points along the domain boundaries. For linear PDEs, we solve the system in one single matrix inversion step. For non-linear PDEs, we apply the Newton-Raphson method. We use a fixed budget of 64 iterations for these non-linear cases. This always guaranties strict convergence. 

We test MetaColloc on six two-dimensional equations. We define the spatial variables as $x$ and $y$. For time-dependent problems, $y$ acts as the time variable $t$. 

The first equation is the Poisson equation. It models smooth spatial changes. We define it as $-\Delta u = f$. The exact solution is $u(x, y) = \sin(2\pi x)\sin(2\pi y) + \exp(-x - y)$. We compute the right-hand side $f$ analytically. 

The second equation is the Helmholtz equation. We configure it as an extremely high-frequency problem with $k = 64\pi$. We define it as $-\Delta u - k^2 u = f$. The exact solution is $u(x, y) = \sin(k_{xy} x)\cos(k_{xy} y) + \exp(-x - y)$, where $k_{xy} = k / \sqrt{2}$. 

The third equation is the Variable Coefficient equation. It models environments with changing physical properties. We define it as $-\nabla \cdot (a \nabla u) = f$. We set the coefficient field as $a(x, y) = 2 + \sin(\pi x)\cos(\pi y)$. The exact solution is $u(x, y) = \sin(\pi x)\sin(\pi y) + \exp(-x - y)$.

The fourth equation is the High-Frequency Poisson equation. We define it identically to the first equation but use an oscillatory solution. The exact solution is $u(x, y) = \sin(8\pi x)\sin(8\pi y) + \exp(-x \cdot y)$. 

The fifth equation is the Sine-Gordon equation. It is a non-linear wave equation. We define it as $u_{yy} - u_{xx} + \sin(u) = f$. The exact solution is $u(x, y) = \sin(\pi x)\cos(\pi y)$. 

The sixth equation is the Korteweg-de Vries (KdV) equation. It models non-linear shallow water waves. It features a difficult third-order derivative. We define it as $u_y + 6u u_x + u_{xxx} = f$. The exact solution is $u(x, y) = \sin(\pi x)\cos(\pi y)$. 

For all six equations, we define the domain as the unit square $[0, 1]^2$. We enforce Dirichlet boundary conditions. We extract the exact boundary values directly from the analytical solutions.

\section{Supplementary Experiments}
\label{app:b}
\subsection{Precision Study}
\label{app:b.1}
We add a study on numerical precision. We want to check if floating-point precision plays a role in the behavior of MetaColloc. We test three settings. The first setting uses full FP32. The second setting uses FP32 for the model and FP64 for the least-squares solve (We denote it as ``FP32+FP64''). The third setting uses full FP64. We run all settings on the same six PDEs. We keep the same seeds and the same hidden sizes. We report the mean RMSE and the 95\% confidence interval.

We show the results in Table~\ref{tab:rmse_pde_comparison_3}. The results show a clear trend. FP32 gives larger errors on several PDEs. FP32+FP64 and full FP64 give almost the same results. This means the least-squares step needs high precision. It also means the model itself does not gain much from full FP64. The gap between FP32 and FP64 does not change the high-frequency behavior. The model still shows the same limits on the high-frequency PDEs. This supports our main claim. The main bottleneck comes from the structure of the basis functions, not from floating-point precision.

\begin{table}[ht]
\centering
\small
\setlength{\tabcolsep}{3pt}
\caption{RMSE comparison (mean with 95\% CI) across six PDEs. Lower is better.}
\label{tab:rmse_pde_comparison_3}
\begin{tabularx}{\textwidth}{l l X X X X X X}
\hline
\textbf{Method} & \textbf{Param} &
\textbf{Poisson} &
\textbf{Helmholtz} &
\textbf{VarCoeff} &
\textbf{HighFreq Poisson} &
\textbf{SineGordon} &
\textbf{KdV} \\
\hline

\multirow{4}{*}{Full FP32}
& H=128  & 1.83e-1 (2.99e-2) & 5.04e-1 (1.73e-3) & 4.83e-3 (4.38e-3) & 8.36e-1 (1.05e-1) & 1.54e-1 (7.25e-2) & 6.05e-1 (1.19e-1) \\
& H=256  & 1.62e-1 (4.96e-3) & 5.26e-1 (6.22e-3) & 1.86e-3 (3.85e-4) & 8.04e-1 (1.24e-1) & 1.05e-1 (1.37e-2) & 9.73e-1 (1.14e-1) \\
& H=512  & 1.68e-1 (7.97e-3) & 5.59e-1 (1.04e-2) & 2.53e-3 (4.21e-4) & 8.00e-1 (1.91e-1) & 1.51e-1 (2.33e-2) & 8.61e-1 (2.30e-1) \\
& H=1024 & 1.70e-1 (7.11e-3) & 6.51e-1 (3.08e-2) & 2.38e-3 (4.14e-4) & \textbf{6.96e-1 (8.30e-2)} & 1.94e-1 (4.44e-2) & 8.95e-1 (2.01e-1) \\
\hline

\multirow{4}{*}{FP32+FP64}
& H=128  & 2.01e-3 (5.12e-4) & 5.02e-1 (1.44e-3) & 1.60e-6 (3.92e-7) & 9.34e-1 (1.02e-1) & 1.09e-4 (5.17e-5) & 9.91e-4 (7.81e-4) \\
& H=256  & \textbf{2.15e-5 (3.15e-6)} & 5.04e-1 (1.49e-3) & 1.87e-9 (9.64e-10) & 1.04e+0 (1.14e-1) & 1.13e-7 (5.33e-8) & 1.37e-4 (4.87e-5) \\
& H=512  & 4.93e-5 (9.68e-6) & 5.12e-1 (3.46e-3) & 2.55e-9 (1.14e-9) & 1.09e+0 (2.15e-1) & 7.62e-8 (2.03e-8) & \textbf{1.20e-4 (4.90e-5)} \\
& H=1024 & 1.07e-4 (2.57e-5) & 5.39e-1 (5.94e-3) & 7.28e-9 (1.12e-9) & 1.36e+0 (4.89e-1) & 5.00e-7 (7.50e-8) & 7.98e-4 (2.34e-4) \\
\hline

\multirow{4}{*}{Full FP64}
& H=128  & 2.47e-3 (3.91e-4) & \textbf{5.02e-1 (1.53e-3)} & 1.83e-6 (4.73e-7) & 1.07e+0 (1.88e-1) & 1.63e-4 (5.78e-5) & 2.49e-3 (9.45e-4) \\
& H=256  & 4.04e-5 (1.19e-5) & 5.03e-1 (1.91e-3) & 1.74e-9 (4.04e-10) & 1.00e+0 (2.06e-1) & 2.79e-7 (1.02e-7) & 1.80e-4 (5.82e-5) \\
& H=512  & 4.23e-5 (1.41e-6) & 5.10e-1 (1.01e-3) & 1.83e-9 (6.99e-10) & 9.64e-1 (1.27e-1) & 9.95e-8 (5.28e-8) & 1.70e-4 (4.93e-5) \\
& H=1024 & 4.28e-5 (1.33e-5) & 5.38e-1 (9.08e-3) & \textbf{1.70e-9 (2.80e-10)} & 9.44e-1 (1.57e-1) & \textbf{5.76e-8 (3.34e-8)} & 1.35e-4 (4.56e-5) \\
\hline
\end{tabularx}
\end{table}

This study shows that FP32 is not enough for stable least-squares solves. FP64 is important for the final solve. But FP64 does not fix the high-frequency limits of the basis. The limits come from the basis itself. This supports our view that the core issue is the mismatch between function approximation and operator behavior.

\subsection{Convergence Behavior of Nonlinear PDE Solvers}
\label{app:b.2}
We also study the behavior of the nonlinear solver. We use the Sine-Gordon equation and the KdV equation. These two PDEs need iterative updates. We use the same Newton--Raphson iteration as in the main text. We test a wide range of iteration counts. We use $K \in \{1,2,4,8,16,32,64,128,256,512\}$, while fixing $H$ at 512. For each setting, We run five seeds and reported the average RMSE value.

\begin{figure}[ht]
    \centering
    \includegraphics[width=0.8\linewidth]{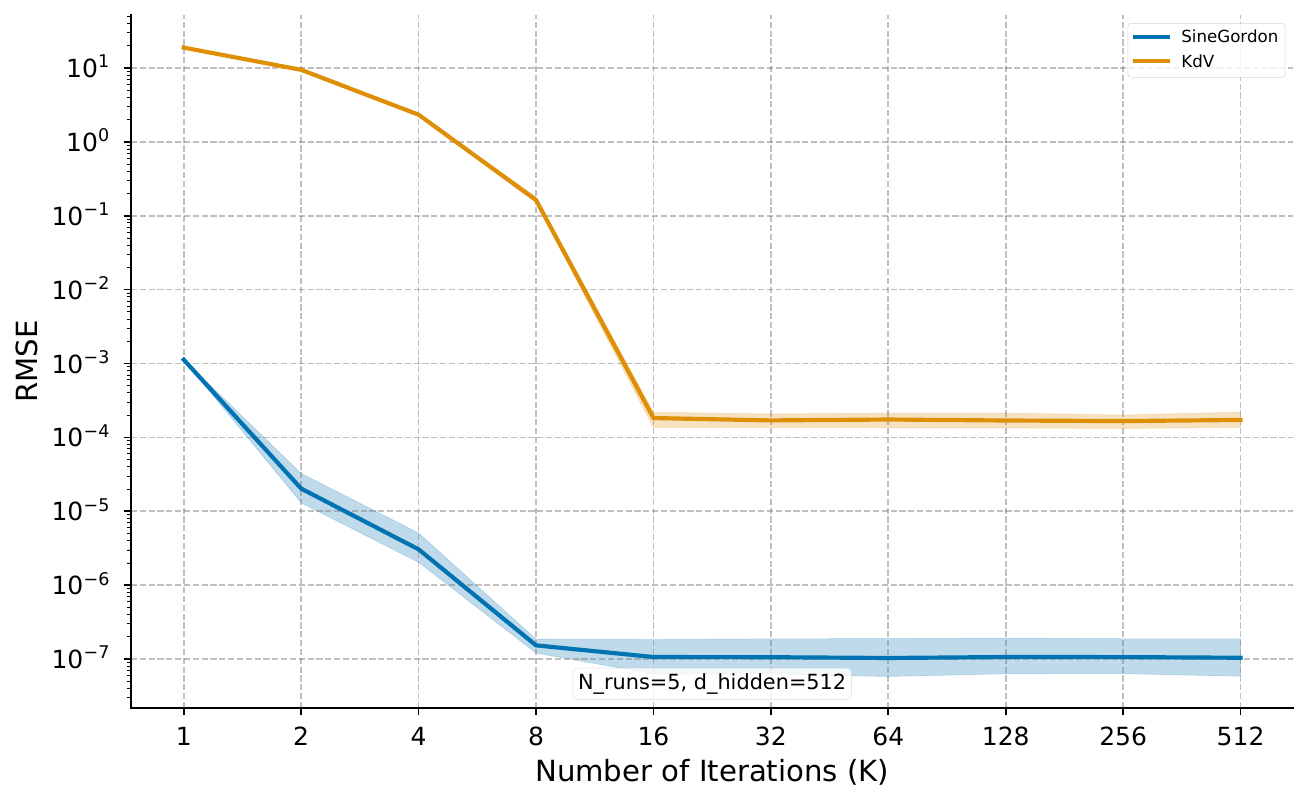}
    \caption{RMSE as a function of the number of iterations $K$ for the Sine-Gordon and KdV equations. We run five seeds for each setting. The curves show stable and smooth decay. The Sine-Gordon equation converges fast. The KdV equation needs more steps but still improves in a steady way.}
    \label{fig:iter_sweep}
\end{figure}

Figure~\ref{fig:iter_sweep} shows the results. The Sine-Gordon equation converges fast. The error drops in the first few steps. The KdV equation needs more steps. The error keeps going down as $K$ grows. Both curves stay stable. Both curves show smooth decay. This means the solver behaves in a predictable way. It also means the choice $K=64$ in the main text is a safe point. It gives a good balance between cost and accuracy.

This study shows that the nonlinear solver is stable. It improves the solution in a steady way. It does not show sudden jumps. It does not show unstable behavior. This supports the main text. The solver is simple but reliable. It works well with the basis learned by MetaColloc.

\subsection{Extension to 3D PDEs}
\label{app:b.3}
We want to prove that MetaColloc easily scales to higher dimensions. Standard grid-based solvers suffer from the curse of dimensionality. Their memory costs explode when moving from 2D to 3D. MetaColloc is completely mesh-free. We only change the input dimension of the neural network from two to three. The core algorithm remains identical.

We train a new MetaColloc model with an input dimension of three. The coordinates are $x, y,$ and $z$. We use the exact same meta-training strategy and hyperparameters described in Appendix~\ref{app:a}. We only adjust the domain to be a three-dimensional unit cube $[0, 1]^3$. During the test phase, we sample 8000 interior points and 3600 boundary points. For non-linear 3D PDEs, we restrict the Newton-Raphson solver to just 8 iterations.

We test this 3D model on three challenging equations. The first is the 3D Poisson equation. It is a standard spatial benchmark. We define it as $u_{xx} + u_{yy} + u_{zz} + f = 0$. The exact solution is $u(x, y, z) = \sin(\pi x)\sin(\pi y)\sin(\pi z)$. 

The second equation is the 3D Burgers equation. We treat $z$ as the time variable. This makes it a 2D spatial plus 1D temporal problem. It models non-linear convection and diffusion. We define it as $u_z + u u_x + u u_y - \nu(u_{xx} + u_{yy}) = f$. We set the viscosity parameter $\nu$ to $0.01$. The exact solution is $u(x, y, z) = \sin(\pi x)\sin(\pi y)\exp(-z)$. 

The third equation is the 3D Allen-Cahn equation. We also treat $z$ as the time variable. It models non-linear reaction and diffusion processes in phase separation. We define it as $u_z - \nu(u_{xx} + u_{yy}) - u(1 - u^2) = f$. We set the diffusion parameter $\nu$ to $0.001$. The exact solution is $u(x, y, z) = \sin(\pi x)\sin(\pi y)\cos(\pi z)$. 

Table \ref{tab:rmse_pde_comparison_3d} shows our results. MetaColloc solves all three equations with remarkable precision. The errors remain firmly in the $10^{-6}$ to $10^{-7}$ range. This proves that our optimization-free neural basis dictionary successfully scales to complex, non-linear spatiotemporal problems without requiring any architectural changes.

\begin{table}[h]
\centering
\small
\setlength{\tabcolsep}{6pt}
\caption{RMSE results for 3D equations. We report the mean and 95\% confidence interval across five seeds.}
\label{tab:rmse_pde_comparison_3d}
\begin{tabularx}{0.8\textwidth}{l X X X}
\hline
\textbf{Param} &
\textbf{Poisson3D} &
\textbf{Burgers3D} &
\textbf{AllenCahn3D} \\
\hline
H=1024  & 3.66e-6 (1.96e-6) & \textbf{1.77e-5 (5.85e-6)} & \textbf{2.24e-6 (1.84e-6)} \\
H=2048  & \textbf{4.04e-7 (6.73e-8)} & 1.95e-5 (5.79e-6) & 3.09e-6 (1.13e-6) \\
\hline
\end{tabularx}
\end{table}

\subsection{Generalization to Complex Geometries and Mixed Boundary Conditions}
\label{app:b.4}
In the main text, our evaluations focus on standard regular domains (e.g., square and cubic boxes) with Dirichlet boundary conditions. A natural question is whether the meta-learned basis dictionary $\Phi_{frozen}$ heavily overfits to Cartesian symmetries or specific boundary types. To demonstrate the true zero-shot geometric flexibility of MetaColloc, we test our pre-trained 2D model (fixed at H=512, seed=42, without any retraining or fine-tuning) on two complex geometric domains with mixed boundary conditions.

\subsubsection{Problem Setup and Boundary Formulations}
Since MetaColloc is a purely point-wise, mesh-free framework, adapting to new geometries only requires sampling collocation points within the new domain. To handle derivative-based boundary conditions, we simply evaluate the directional derivatives of our neural basis using forward-mode automatic differentiation.

For a boundary point $x_{bd}$ with outward normal vector $n$, the Neumann boundary condition requires $\nabla u \cdot n = h$. We assemble the corresponding row in the boundary matrix $A_{bd}$ as: $$A_{bd} = \Phi_x n_x + \Phi_y n_y$$ Similarly, for a Robin boundary condition $u + \alpha \nabla u \cdot n = g_r$, the matrix row is assembled as: $$A_{bd} = \Phi + \alpha (\Phi_x n_x + \Phi_y n_y)$$ We solve the Poisson equation $-\Delta u = f$ with the exact solution $u(x,y) = \sin(2\pi x)\sin(2\pi y) + \exp(-x-y)$ on the following two irregular domains.

\paragraph{L-Shape Domain with Mixed Dirichlet-Neumann BCs:}
The domain is defined as $[-1, 1]^2 \setminus [0, 1] \times [-1, 0]$, featuring a challenging re-entrant corner at the origin. We apply Dirichlet conditions on the outer left, top, and bottom-left edges. We apply Neumann conditions (matching the exact normal derivatives) on the right-top edge and the two inner edges forming the concave corner.

\paragraph{Annulus Domain with Mixed Dirichlet-Robin BCs:}
The domain is a ring defined by $r \in [0.5, 1.0]$. We apply a Dirichlet condition on the inner boundary ($r=0.5$) and a Robin condition (with $\alpha=1$) on the outer boundary ($r=1.0$).

\subsubsection{Results and Discussion}
We sampled 3000 interior collocation points and 600 boundary points for both geometries. The linear system was solved in a single step using double precision (FP64).

\paragraph{L-Shape (Dirichlet + Neumann):} The model achieved an RMSE of 2.00e-03.
\begin{figure}[ht]
    \centering
    \includegraphics[width=\linewidth]{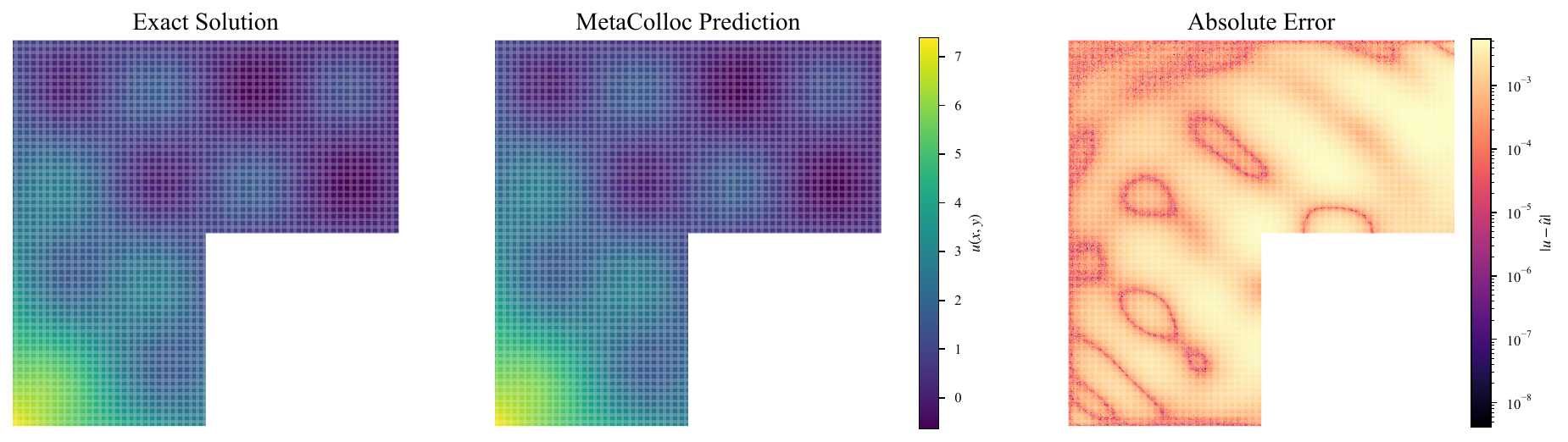}
    \caption{Exact solution, MetaColloc prediction, and absolute error heatmap for the Poisson equation on L-shape domain with mixed Dirichlet and Neumann boundaries. The error is well-bounded across the entire domain, including the challenging re-entrant corner of the L-shape.}
    \label{fig:lshape}
\end{figure}

\paragraph{Annulus (Dirichlet + Robin):} The model achieved an RMSE of 3.48e-04.
\begin{figure}[ht]
    \centering
    \includegraphics[width=\linewidth]{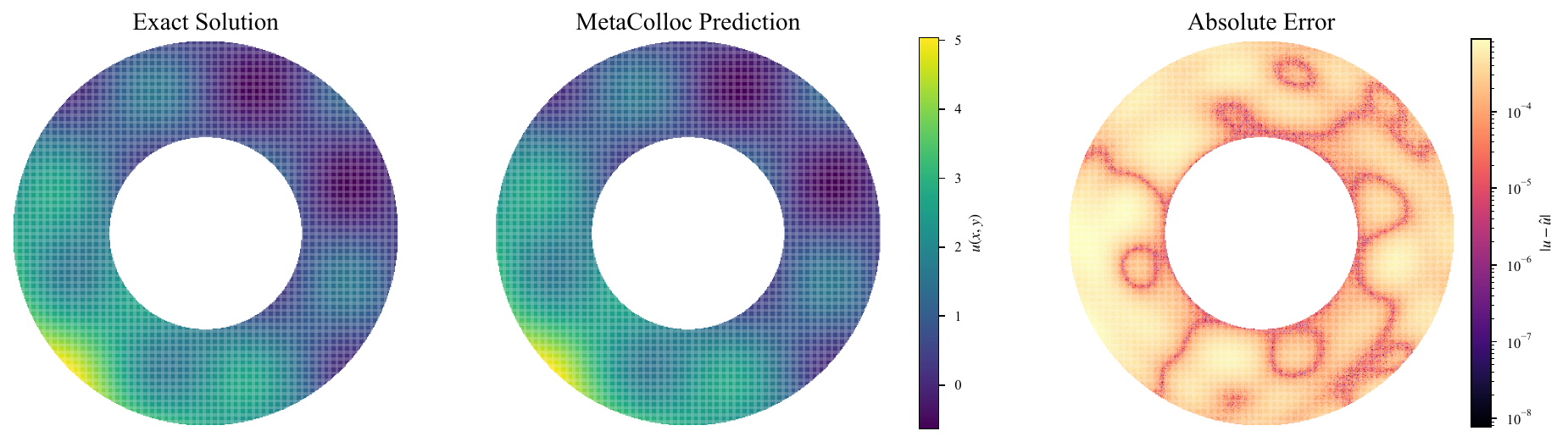}
    \caption{Exact solution, MetaColloc prediction, and absolute error heatmap for the Poisson equation on Annulus domain with mixed Dirichlet and Robin boundaries. The error is well-bounded across the entire domain.}
    \label{fig:annulus}
\end{figure}

These results explicitly confirm that the neural basis functions learned via multi-scale Gaussian Random Fields are highly expressive and independent of the spatial grid. MetaColloc can seamlessly assemble collocation matrices for complex geometries and arbitrary linear boundary conditions out-of-the-box, retaining its extreme test-time efficiency.

\subsection{Time Usage in Section~\ref{sec:4.2}}
\label{app:b.5}
We first show the wall-clock time required for MetaColloc training, as shown in the table~\ref{tab:metacolloc_training_time}

\begin{table}[ht]
\centering
\small
\setlength{\tabcolsep}{3pt}
\caption{MetaColloc training time (mean with 95\% CI) for different $H$.}
\label{tab:metacolloc_training_time}
\begin{tabularx}{0.28\textwidth}{l l}
\hline
\textbf{Param} & \textbf{Training Time (s)} \\
\hline
H=128  & 403.46 (3.00) \\
H=256  & 468.18 (1.32) \\
H=512  & 701.22 (0.59) \\
H=1024 & 1414.86 (1.07) \\
\hline
\end{tabularx}
\end{table}

We report the wall time consumed in solving each equation in the experiments in Section~\ref{sec:4.2} as shown in the table~\ref{tab:time_pde_comparison}. We notice that MetaColloc consumes an extremely short time, which is our core advantage. It is used to solve equations only after pre-training, and solving the final $w$ is very fast on the GPU.

\begin{table}[ht]
\centering
\small
\setlength{\tabcolsep}{3pt}
\caption{Solving time comparison (mean with 95\% CI) across six PDEs. The unit of data is seconds. Lower is better.}
\label{tab:time_pde_comparison}
\begin{tabularx}{\textwidth}{l l X X X X X X}
\hline
\textbf{Method} & \textbf{Param} &
\textbf{Poisson} &
\textbf{Helmholtz} &
\textbf{VarCoeff} &
\textbf{HighFreq Poisson} &
\textbf{SineGordon} &
\textbf{KdV} \\
\hline

\multirow{4}{*}{MetaColloc}
& H=128  & 1.590 (0.449) & 1.306 (0.027) & 1.287 (0.002) & 1.287 (0.003) & 1.461 (0.003) & 15.981 (0.046) \\
& H=256  & 1.298 (0.008) & 1.292 (0.005) & 1.292 (0.005) & 1.291 (0.004) & 1.628 (0.010) & 16.153 (0.063) \\
& H=512  & 1.430 (0.097) & 1.304 (0.003) & 1.304 (0.004) & 1.304 (0.003) & 1.862 (0.003) & 16.459 (0.030) \\
& H=1024 & 1.543 (0.261) & 1.296 (0.013) & 1.295 (0.011) & 1.295 (0.011) & 2.331 (0.020) & 16.746 (0.143) \\
\hline

\multirow{4}{*}{PINN L-BFGS}
& H=128  & 73.007 (0.764) & 73.007 (0.764) & 73.007 (0.764) & 73.007 (0.764) & 73.007 (0.764) & 73.007 (0.764) \\
& H=256  & 76.570 (0.484) & 76.570 (0.484) & 76.570 (0.484) & 76.570 (0.484) & 76.570 (0.484) & 76.570 (0.484) \\
& H=512  & 95.046 (1.133) & 95.046 (1.133) & 95.046 (1.133) & 95.046 (1.133) & 95.046 (1.133) & 95.046 (1.133) \\
& H=1024 & 159.933 (0.679) & 159.933 (0.679) & 159.933 (0.679) & 159.933 (0.679) & 159.933 (0.679) & 159.933 (0.679) \\
\hline

GP-HM & Grid=100$^2$ & 4643.387 (3.303) & 4676.184 (1.470) & 5053.207 (2.532) & 4869.284 (215.580) & 432.252 (0.453) & 1085.075 (0.155) \\
\hline

\multirow{4}{*}{ConFIG}
& H=128  & 136.707 (0.051) & 136.707 (0.051) & 136.707 (0.051) & 136.707 (0.051) & 136.707 (0.051) & 136.707 (0.051) \\
& H=256  & 140.095 (0.376) & 140.095 (0.376) & 140.095 (0.376) & 140.095 (0.376) & 140.095 (0.376) & 140.095 (0.376) \\
& H=512  & 210.545 (3.953) & 210.545 (3.953) & 210.545 (3.953) & 210.545 (3.953) & 210.545 (3.953) & 210.545 (3.953) \\
& H=1024 & 473.149 (4.445) & 473.149 (4.445) & 473.149 (4.445) & 473.149 (4.445) & 473.149 (4.445) & 473.149 (4.445) \\
\hline

\multirow{4}{*}{PINN DCGD}
& H=128  & 187.939 (3.544) & 187.939 (3.544) & 187.939 (3.544) & 187.939 (3.544) & 187.939 (3.544) & 187.939 (3.544) \\
& H=256  & 184.260 (0.864) & 184.260 (0.864) & 184.260 (0.864) & 184.260 (0.864) & 184.260 (0.864) & 184.260 (0.864) \\
& H=512  & 252.974 (0.675) & 252.974 (0.675) & 252.974 (0.675) & 252.974 (0.675) & 252.974 (0.675) & 252.974 (0.675) \\
& H=1024 & 473.192 (8.529) & 473.192 (8.529) & 473.192 (8.529) & 473.192 (8.529) & 473.192 (8.529) & 473.192 (8.529) \\
\hline
\end{tabularx}
\end{table}

\section{Gaussian Random Field Sampling}
\label{app:c}
We describe the sampling process used in our meta-training stage. We use Gaussian Random Fields (GRFs) to generate a wide range of target functions. These functions cover smooth shapes and high-frequency waves. This diversity is important. It forces the network to learn a flexible basis.

We sample functions in three modes. We choose the mode at random for each task. The modes are:
\begin{itemize}
    \item \textbf{RBF mode}: smooth functions with a random length scale.
    \item \textbf{High-frequency mode}: oscillatory functions with a random center frequency and bandwidth.
    \item \textbf{Mixed mode}: a combination of the two.
\end{itemize}

We draw $M = n_{\text{train}} + n_{\text{test}}$ input points $X \in [0,1]^2$ from a uniform distribution. We then build a random feature map

\[
\phi(x) = \sqrt{\frac{2}{D}} \cos(\omega^\top x + b),
\]

where $D$ is the number of random features. The vector $b$ contains random phases drawn from a uniform distribution on $[0, 2\pi]$. The vector $\omega$ depends on the sampling mode.

\paragraph{RBF mode.}
We draw a log-uniform length scale $\ell \in [\ell_{\text{lo}}, \ell_{\text{hi}}]$. We then sample

\[
\omega \sim \mathcal{N}(0, \ell^{-2} I_2).
\]

This produces smooth functions with controlled variation.

\paragraph{High-frequency mode.}
We draw a center frequency $\mu$ from a uniform range $[\mu_{\text{lo}}, \mu_{\text{hi}}]$. We draw a bandwidth $\sigma$ from $[\sigma_{\text{lo}}, \sigma_{\text{hi}}]$. We sample half of the frequencies from

\[
\omega \sim \mathcal{N}(\mu, \sigma^2 I_2),
\]

and the other half from its symmetric counterpart $-\omega$. This creates oscillatory functions with balanced positive and negative frequencies.

\paragraph{Mixed mode.}
We split the features into two halves. One half uses the RBF mode. The other half uses the high-frequency mode.

\paragraph{Output construction.}
We draw random weights $w \sim \mathcal{N}(0, I_D)$. We compute the target values

\[
y = \phi(X) w.
\]

We split the first $n_{\text{train}}$ samples for training and the rest for testing.

This sampling process gives a broad and controlled distribution of functions. It covers smooth gradients, sharp transitions, and high-frequency waves. This diversity is essential for learning a universal basis that generalizes to many PDEs.


\end{document}